\documentclass{article}

\usepackage{graphicx}
\usepackage{subfigure}
\usepackage{booktabs} 
\usepackage{amsmath}
\usepackage{amssymb}
\usepackage{censor}

\graphicspath{{./figs/}}

\usepackage{hyperref}

\newif\ifanon

\anonfalse

\ifanon
\usepackage{icml2020}
\else
\usepackage[accepted]{icml2020}
\fi
\icmltitlerunning{Graph Prolongation Convolutional Networks}

\DeclareMathOperator{\Tr}{Tr}

\begin{document}

\twocolumn[
\icmltitle{Graph Prolongation Convolutional Networks: \\
           Explicitly Multiscale Machine Learning on Graphs \\
           with Applications to Modeling of Cytoskeleton
           }

\icmlsetsymbol{equal}{*}

\begin{icmlauthorlist}
\icmlauthor{Cory B. Scott}{uci}
\icmlauthor{Eric Mjolsness}{uci}
\end{icmlauthorlist}

\icmlaffiliation{uci}{Department of Computer Science, University of California Irvine, Irvine, California, USA}

\icmlcorrespondingauthor{Cory B. Scott}{scottcb@uci.edu}

\icmlkeywords{Machine Learning, Graph Convolution, }

\vskip 0.3in
]

\printAffiliationsAndNotice{} 
\thispagestyle{plain}
\pagestyle{plain}
\setlength{\footskip}{20pt}
\begin{abstract}
We define a novel type of ensemble Graph Convolutional Network (GCN) model. Using optimized linear projection operators to map between spatial scales of graph, this ensemble model learns to aggregate information from each scale for its final prediction. We calculate these linear projection operators as the infima of an objective function relating the structure matrices used for each GCN. Equipped with these projections, our model (a Graph Prolongation-Convolutional Network) outperforms other GCN ensemble models at predicting the potential energy of monomer subunits in a coarse-grained mechanochemical simulation of microtubule bending. We demonstrate these performance gains by measuring an estimate of the FLOPs spent to train each model, as well as wall-clock time. Because our model learns at multiple scales, it is possible to train at each scale according to a predetermined schedule of coarse vs. fine training. We examine several such schedules adapted from the Algebraic Multigrid (AMG) literature, and quantify the computational benefit of each. We also compare this model to another model which features an optimized coarsening of the input graph. Finally, we derive backpropagation rules for the input of our network model with respect to its output, and discuss how our method may be extended to very large graphs.  
\end{abstract}

\section{Introduction}
\label{sec:intro}
\subsection{Convolution and Graph Convolution}
Recent successes of deep learning have demonstrated that the inductive bias of Convolutional Neural Networks (CNNs) makes them extremely efficient for analyzing data with an inherent grid structure, such as images or video.
In particular, many applications use these models to make per-node (per-pixel) predictions over grid graphs: examples include image segmentation, optical flow prediction, anticipating motion of objects in a scene, and facial detection/identification.
Further work applies these methods to emulate physical models, by discretizing the input domain. Computational Fluid Dynamics and other scientific tasks featuring PDEs or ODEs on a domain discretized by a rectangular lattice have seen recent breakthroughs applying machine learning models, like CNNs to handle data which is structured this way. These models learn a set of local filters whose size is much smaller than the size of the domain - these filters may then be applied simultaneously across the entire domain, leveraging the fact that at a given scale the local behavior of the neighborhood around a pixel (voxel) is likely to be similar at all grid points. 

Graph Convolutional Networks (GCNs) are a natural extension of the above idea of image `filters' to arbitrary graphs rather than $n$D grids, which may be more suitable in some scientific contexts. Intuitively, GCNs replace the image filtering operation of CNNs with repeated passes of: 1) aggregation of information between nodes according to some structure matrix 2) nonlinear processing of data at each node according to some rule (most commonly a flat neural network which takes as separate input(s) the current vector at each node). We refer the reader to a recent survey by Bacciu et al \yrcite{bacciu2019gentle} for a more complete exploration of the taxonomy graph neural networks. 

\subsection{Microtubules}
As an example of a dataset whose underlying graph is not a grid, we consider a coarse-grained simulation of a microtubule. Microtubules (MTs) are self-assembling nanostructures,
ubiquitous in living cells, that along with actin filaments
comprise a major portion of the dynamic cytoskeleton
governing cell shape and mechanics.
Whole-MT biomechanical models would be a useful tool
for modeling cytoskeletal dynamics at the cellular scale. Microtubules play important structural roles during cell division, cell growth, and separation of chromosomes (in eukaryotic cells) \cite{chakrabortty2018computational}. Microtubules are comprised of a lattice structure of two conformations ($\alpha$ and $\beta$) of tubulin. Free-floating tubulin monomers associate energetically into dimer subunits, which then associate head-to-tail to form long chain-like complexes called \emph{protofilaments}. Protofilaments associate side-to side in a sheet; at some critical number of protofilaments (which varies between species and cell type) the sheet wraps closed to form a repeating helical lattice with a seam. See \cite{pampaloni2008microtubule}, Page 303, Figure 1. Key properties of microtubules are: \\
{\bf Dynamic instability:} microtubules grow from one end by attracting free-floating tubulin monomers  \cite{vanburen2005mechanochemical}. Microtubules can spontaneously enter a ``catastrophe'' phase, in which they rapidly unravel, but can also ``rescue'' themselves from the catastrophe state and resume growth \cite{gardner2013microtubule, shaw2003sustained}. \\
{\bf Interactions:} Microtubules interact with one another: they can dynamically avoid one another during the growth phase, or collide and bundle up, or collide and enter catastrophe \cite{tindemans2014efficient}. The exact mechanism governing these interactions is an area of current research. \\
{\bf Structural strength:} microtubules are very stiff, with a Young's Modulus estimated at $\approx$1GPa for some cases \cite{pampaloni2008microtubule}. This stiffness is thought to play a role in reinforcing cell walls \cite{kis2002nanomechanics}. 

In this work we introduce a model which learns to reproduce the dynamics of a graph signal (defined as an association of each node in the network with a vector of discrete or real-valued labels) at multiple scales of graph resolution. We apply this model framework to predict the potential energy of each tubulin monomer in a mechanochemical simulation of a microtubule.

\subsection{Simulation of MTs and Prior Work}
Non-continuum, non-event-based simulation of large molecules is typically done by representing some molecular subunit as a particle/rigid body, and then defining rules for how these subunits interact energetically. Molecular Dynamics (MD) simulation is an expansive area of study and a detailed overview is beyond the scope of this paper. We instead proceed to describe in general terms some basic ideas relevant to the numerical simulation detailed in Section \ref{subsec:data_gen}. 
MD simulations proceed from initial conditions by computing the forces acting on each particle (according to the potential energy interactions and any external forces, as required), determining their instantaneous velocities and acceleration accordingly, and then moving each particle by the distance it would move (given its velocity) for some small timestep. Many variations of this basic idea exist. The software we use for our MD simulations, LAMMPS \cite{plimpton1993fast} allows for many different types of update step: we use Verlet integration (updating particle position according to the central difference approximation of acceleration \cite{verlet1967computer}) and Langevin dynamics (modeling the behavior of a viscous surrounding solvent implicitly \cite{schneider1978molecular}).  We also elect to use the microcanonical ensemble (NVE) - meaning that the update steps of the system maintain the total number of particles, the volume of the system, and the total energy (kinetic + potential). For more details of our simulation, see Section \ref{subsec:data_gen} and the source code, available in the Supplementary Material accompanying this paper. Independent of implementation details, a common component of many experiments in computational molecular dynamics is the prediction of the potential energy associated with a particular conformation of some molecular structure. Understanding the energetic behavior of a complex molecule yields insights into its macro-scale behavior: for instance, the problem of protein folding can be understood as seeking a lower-energy configuration. In this work, we apply graph convolutional networks, trained via a method we introduce, to predict these energy values for a section of microtubule.

\subsection{Mathematical Background and Notation}
\label{sec:prior}
\paragraph{Definitions:} For all basic terms (graph, edge, vertex, degree) we use standard definitions. We use the notation ${\left\{x_i \right\}}_{i=a}^{b}$ to represent the sequence of $x_i$ indexed by the integers $a, a+1, a+2, \ldots b$. When $X$ is a matrix, we will write $\left[ X \right]_{ij}$ to denote the entry in the $i$th row, $j$th column.\\
{\bf Graph Laplacian:} The graph Laplacian is the matrix given by $L(G) = A(G) - \text{diag}(A(G) \cdot \mathbf{1})$ where $A(G)$ is the adjacency matrix of $G$, and $\mathbf{1}$ is an appropriately sized vector of 1s. The graph Laplacian is given by some authors as the opposite sign. \\
{\bf Linear Graph Diffusion Distance (GDD):} Given two graphs $G_1$ and $G_2$, with $|G_1| \leq |G_2|$ the Linear Graph Diffusion Distance $D(G_1, G_2)$ is given by:
    \begin{align}
         D(G_1, G_2) = \inf_{\substack{P|\mathcal{C}(P)\\\alpha > 0}} {\left| \left| \frac{1}{\alpha} P L(G_1) - \alpha L(G_2) P \right| \right|}_F \label{eqn:GDD}
    \end{align}
    where $\mathcal{C}(P)$ represents some set of constraints on $P$, $\alpha$ is a scalar with $\alpha > 0$, and $||\cdot||_F$ represents the Frobenius norm.   We take $\mathcal{C}(P)$ to be orthogonality: $P^T P = I$. Note that since in general $P$ is a rectangular matrix, it may not be the case that $P P^T = I$. Unless stated otherwise all $P$ matrices detailed in this work were calculated with $\alpha = 1$, using the procedure laid out in the following section, in which we briefly detail an algorithm for efficiently computing the distance in the case where $\alpha$ is allowed to vary. The efficiency of this algorithm is necessary to enable the computation of the LGDD between very large graphs, as discussed in Section \ref{subsec:graph_limits}. \\
{\bf Prolongation matrix:} we use the term ``prolongation matrix'' to refer to a matrix which is the optimum of the minimization given in the definition of the LGDD. 

\subsection{Efficient Calculation of Graph Diffusion Distance}
\begin{figure}[h]
    \centering
    \includegraphics[width=\linewidth]{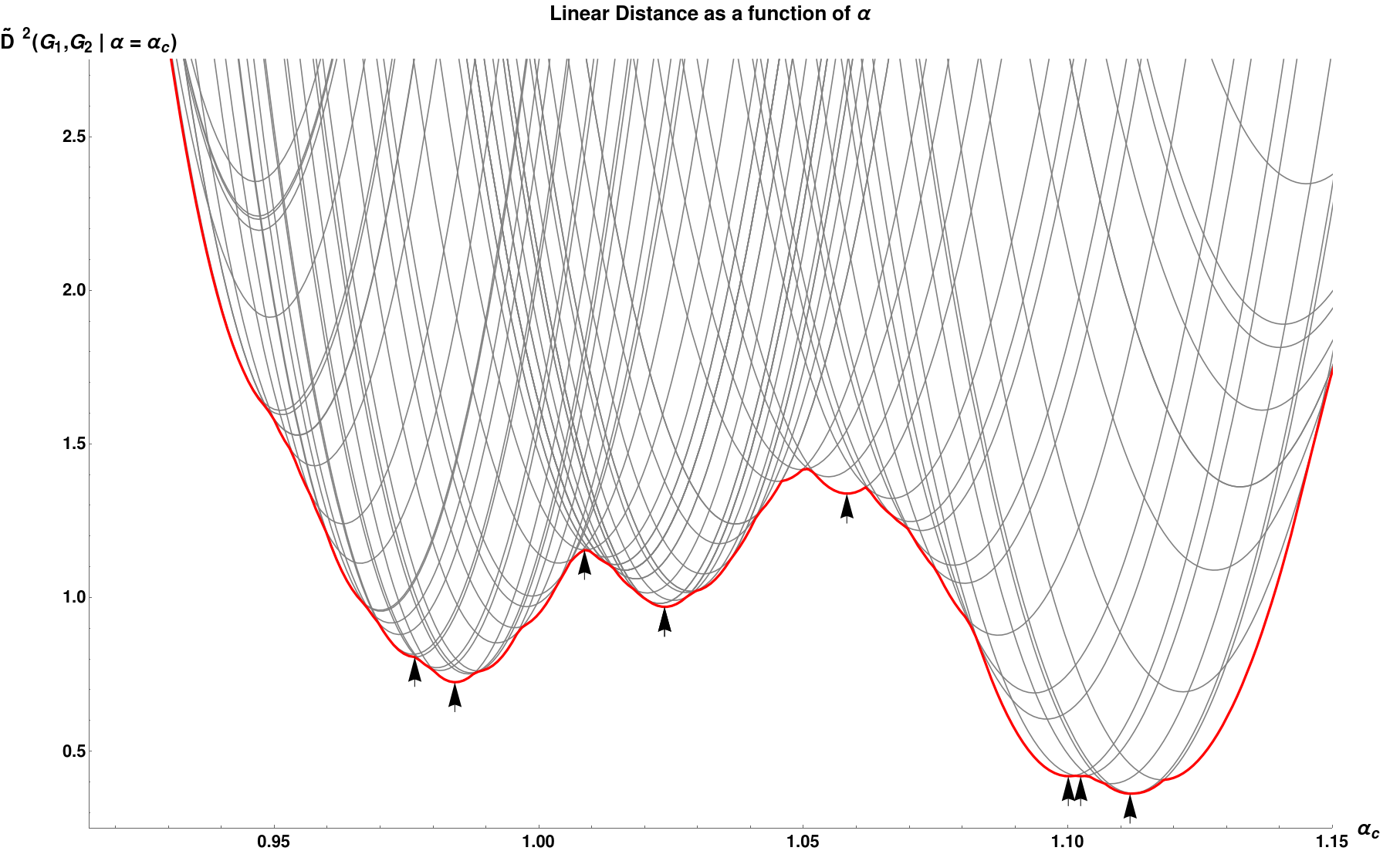}
    \caption{Plot of Linear Graph Diffusion Distance between two small random graphs, as $\alpha$ is varied. Each grey curve shows the objective function when $P$ is fixed, as a function of $\alpha$, and each curve represents a $P$ matrix which is optimal at any value of $\alpha$ in the plotted range. The red curve shows the lower convex hull of all grey curves. Note that it is continuous but has discontinuous slope. Black arrows represent local optima. The discontinuous slope and high number of local optima illustrate why optimizing this function using univariate search over $\alpha$ is inefficient.}
    \label{fig:lin_dist_ex}
\end{figure}

The joint optimization given in the definition of Linear Graph Diffusion Distance (Equation \ref{eqn:GDD}) is a nested optimization problem. If we set
\begin{align*}
    f(\alpha) &= D(G_1, G_2 | \alpha) \\
              &= \inf_{P|\mathcal{C}(P)} {\left| \left| \frac{1}{\alpha} P L(G_1) - \alpha L(G_2) P \right| \right|}_F,
\end{align*}
then each evaluation of $f$ requires a full optimization of the matrix $P$ subject to constraints $\mathcal{C}$. When $L(G_1)$ and $L(G_2)$ are Graph Laplacians, $f(\alpha)$ is continuous, but with discontinuous derivative, and has many local minima (see Figure \ref{fig:lin_dist_ex}). As a result, the naive approach of optimizing $f(\alpha)$ using a univariate optimization method like Golden Section Search is inefficient. In this section we briefly describe a procedure for performing this joint optimization more efficiently. For a discussion of variants of the LGDD, as well as the theoretical justification of this algorithm, see \cite{scott2019novel}. 

First, we note that by making the constraints on $P$ more restrictive, we upper-bound the original distance:
    \begin{align}
         D(G_1, G_2) &= \inf_{\substack{P|\mathcal{C}(P)\\\alpha > 0}} {\left| \left| \frac{1}{\alpha} P L(G_1) - \alpha L(G_2) P \right| \right|}_F \nonumber \\
         &\leq \inf_{\substack{P|\mathcal{S}(P)\\\alpha > 0}} {\left| \left| \frac{1}{\alpha} P L(G_1) - \alpha L(G_2) P \right| \right|}_F. \label{eqn:GDD2}
    \end{align}
In our case, $\mathcal{C}(P)$ represents orthogonality. As a restriction of our constraints we specify that $P$ must be related to a \emph{subpermutation}  matrix (an orthogonal matrix having only 0 and 1 entries) $\Tilde{P}$ as follows: $ P = U_2 \Tilde{P} U_1^T$, where the $U_i$ are the fixed matrices which diagonalize $L(G_i)$: $L(G_i) = U_i \Lambda_i U_i^T$. Then, 

    \begin{align}
         D(G_1, G_2) &\leq \inf_{\substack{P|\mathcal{S}(P)\\\alpha > 0}} {\left| \left| \frac{1}{\alpha} P L(G_1) - \alpha L(G_2) P \right| \right|}_F \nonumber \\
         &= \inf_{\substack{\Tilde{P}|\text{subperm}(\Tilde{P})\\\alpha > 0}} {\left| \left| \frac{1}{\alpha} U_2 \Tilde{P} U_1^T U_1 \Lambda_1 U_1^T \right. \right.} \nonumber \\
         & \null \qquad \qquad \qquad {\left. \left. - \alpha U_2 \Lambda_2 U_2^T U_2 \Tilde{P} U_1^T \right| \right|}_F \nonumber \\
         &= \inf_{\substack{\Tilde{P}|\text{subperm}(\Tilde{P})\\\alpha > 0}} {\left| \left| \frac{1}{\alpha} U_2 \Tilde{P} \Lambda_1 U_1^T  - \alpha U_2 \Lambda_2 \Tilde{P} U_1^T \right| \right|}_F \nonumber \\
         &= \inf_{\substack{\Tilde{P}|\text{subperm}(\Tilde{P})\\\alpha > 0}} {\left| \left| U_2 \left( \frac{1}{\alpha} \Tilde{P} \Lambda_1 - \alpha \Lambda_2 \Tilde{P} \right) U_1^T \right| \right|}_F. \nonumber 
         \intertext{Because the $U_i$ are rotation matrices (under which the Frobenius norm is invariant), this further simplifies to}
         D(G_1, G_2) &\leq \inf_{\substack{\Tilde{P}|\text{subperm}(\Tilde{P})\\\alpha > 0}} {\left| \left| \frac{1}{\alpha} \Tilde{P} \Lambda_1 - \alpha \Lambda_2 \Tilde{P} \right| \right|}_F. \nonumber 
    \end{align}
Furthermore, because the $\Lambda_i$ are diagonal, this optimization is equivalent to a Rectangular Linear Assignment Problem (RLAP) \cite{bijsterbosch2010solving}, between the diagonal entries $\lambda^(1)_j$ and $\lambda^(2)_l$ of $\Lambda_1$ and $\Lambda_2$, respectively, with the $\alpha$-dependent cost of an assignment given by:
\begin{align}
    c_\alpha(\lambda^(1)_j, \lambda^(2)_l) = {\left( \frac{1}{\alpha} \lambda^(1)_j - \alpha \lambda^(2)_l \right) }^2.
\end{align}.

RLAPs are extensively studied. We use the general LAP solving package lapsolver \cite{heindl2018lapsolver} to comute $\Tilde{P}$. In practice (and indeed in this paper) we set often set $\alpha = 1$, in which case the solution $\Tilde{P}$ of the RLAP only acts as a preconditioner for the orthogonally-constrained optimization over $P$. More generally, when alpha is allowed to vary (and therefore many RLAPs must be solved), a further speedup is attained by re-using partial RLAP solutions from previously-tested values of $\alpha$ to find the optimal assignment at $\alpha'$. We detail how this may be done in out recent work \cite{scott2019novel}. 

For the $P$ matrices used in the experiments in this work, we set $\alpha=1$ and used lapsolver to find an optimal assignment $\Tilde{P}$. We then initialized an orthogonally-constrained optimization of \ref{eqn:GDD} with $ P = U_2 \Tilde{P} U_1^T$. This constrained optimization was performed using Pymanopt \cite{townsend2016pymanopt}. 

\section{Model Architecture}
\label{sec:model_info}
The model we propose is an ensemble of GCNs at multiple scales, with optimized projection matrices performing the mapping in between scales (i.e. between ensemble members).
More formally, Let ${\left\{G_i \right\}}_{i=1}^{k}$ represent a sequence of graphs with $\left| G_1 \right| \geq \left| G_2 \right| \ldots \geq \left| G_k \right| $, and let ${\left\{Z_i = z(G_i) \right\}}_{i=1}^{k}$ be their structure matrices (for some chosen method $z$ of calculating the structure matrix given the graph). In all experiments in this paper, we take $z(G) = L(G)$, the graph Laplacian, as previously defined \footnote{Other GCN research uses powers of the Laplcian, the normalized Laplacian, the symmetric normalized laplacian, etc. Comparison of these structure matrices is out of scope of this paper.}. In an ensemble of Graph Convolutional Networks, let $\theta^{(i)}_l = \left\{ W^{(i)}_l, b^{(i)}_l \right\}$ represent the parameters (filter matrix and bias vector) in layer $l$ of the $i$th network.

We follow the GCN formulation given by Kipf and Welling \yrcite{kipf2016semi}. Assuming an input tensor $X$ of dimensions $n \times F$ (where $n$ is the number of nodes in the graph and $F$ is the dimension of the label at each node), we inductively define the layerwise update rules for a graph convolutional network $\textsc{gcn}\left(Z_i, X, {\left\{\theta^{(i)}_{l} \right\}}_{l=1}^{m}\right)$ as:
     \begin{align*}
        X_0 &= X \\
        X_m &= g_m\left(Z_i X_{m-1} W^{(i)}_m +b^{(i)}_m \right),
     \end{align*}
where $g_m$ is the activation function of the $m$th layer.  

When $i=j-1$, let $P_{i,j}$ be an optimal (in either the sense of Graph Diffusion Distance, or in the sense we detail in section \ref{subsec:compare_gcn}) prolongation matrix from $L(G_j)$ to $L(G_i)$, i.e. $
    P_{i,j} = \arg \inf_{P|\mathcal{C}(P)} {\left| \left| P L(G_j) - L(G_i) P \right| \right|}_F.$
Then, for $i < j-1$, let $P_{i,j}$ be shorthand for the matrix product $P_{i,i+1} P_{i+1,i+2} \ldots P_{j-1,j}$. For example, $P_{1,4} = P_{1,2}P_{2,3}P_{3,4}$.

\begin{figure}[h]
        \centering
        \includegraphics[width = \linewidth]{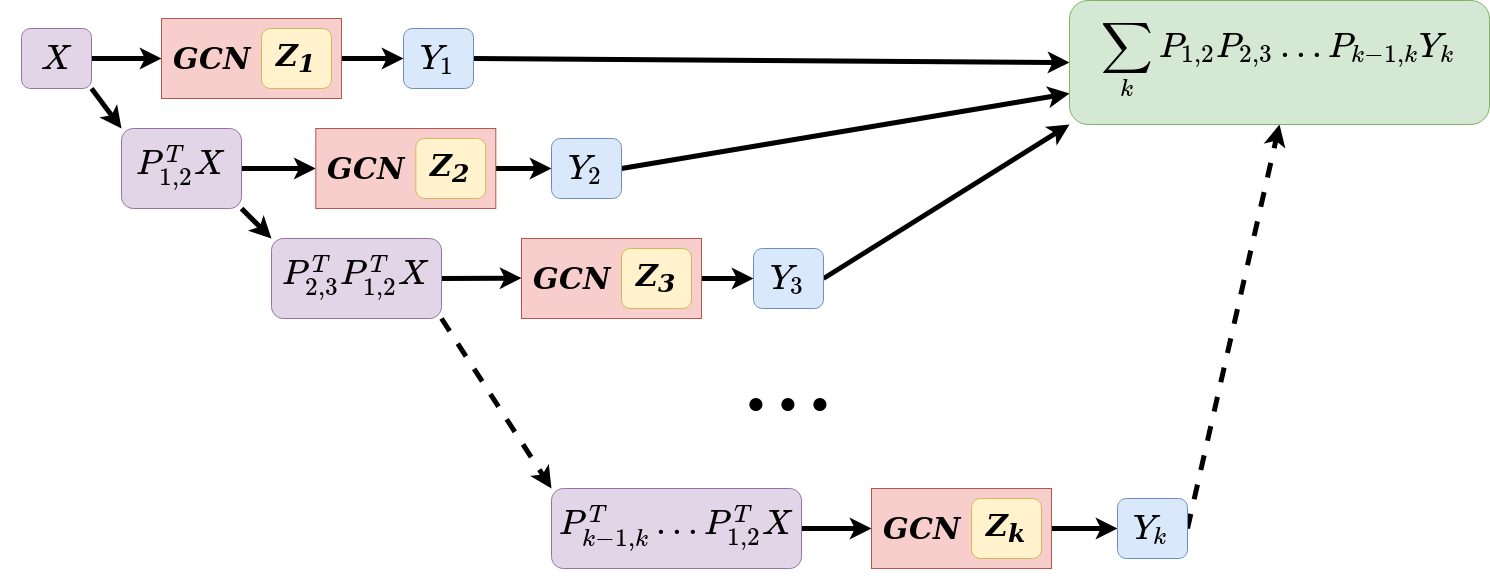}
        \caption{Schematic of GPCN model. Data matrix $X$ is fed into the model and repeatedly coarsened using optimized projection matrices $P_{ik}$. These coarsened data matrices are separately fed into GCN models. The final output of the ensemble is the projected sum of the outputs of each component GCN.}
        \label{fig:gcn_schematic}
    \end{figure}

Our multiscale ensemble model is then constructed as: \\
    \begin{align}
     &\textbf{GPCN}\left(
        \left\{Z_{i} \right\}_{i=1}^{k},
        X,
        \left\{ \left\{\theta^{(i)}_l \right\}_{l=1}^{m_i}  \right\}_{i=1}^{k},
        \left\{ P_{i,i+1} \right\}_{i=1}^{k-1}
    \right) \nonumber\\
    &\null\quad = \textsc{gcn}\left(Z_1, X, \left\{\theta^{(1)}_l \right\}_{l=1}^{m_1} \right)\nonumber\\
    &\null\quad\quad+ \sum_{i=2}^{k} P_{1i}  \textsc{gcn}\left(
        Z_i, 
        P_{1i}^T X,
        \left\{\theta^{(i)}_l \right\}_{l=1}^{m_i}
    \right)     \label{eqn:gpcn_forward_rule}
    \end{align}
    This model architecture is illustrated in Figure \ref{fig:gcn_schematic}. When the $P$ matrices are constant/fixed, we will refer to this model as a GPCN, for Graph Prolongation-Convolutional Network. However, we find in our experiments in Section \ref{subsec:compare_gcn} that validation error is further reduced when the $P$ operators are tuned during the same gradient update step which updates the filter weights, which we refer to as an ``adaptive'' GPCN or A-GPCN. We explain our method for choosing $Z_i$ and optimizing $P$ matrices in Section \ref{subsec:compare_gcn}.

\section{Numerical Experiments}
\label{sec:exp}

\subsection{Dataset}
\label{subsec:data_gen}
In this Section we detail the process for generating the simulated microtubule data for comparison of our model with other GCN ensemble models. Our microtubule structure has 13 protofilaments (each 48 tubulin monomers long). As in a biological microtubule, each tubulin monomer is offset (along the axis parallel to the protofilaments) from its neighbors in adjacent protofilaments, resulting in a helical structrure with a pitch of 3 tubulin units. We refer to this pitch as the ``offset'' in Section \ref{subsec:coarse}. Each monomer subunit (624 total) is represented as a point mass of 50 Dalton ($8.30 \times 10^{-15}$ng). The diameter of the whole structure is 26nm, and the length is $\approx260$nm. The model itself was constructed using Moltemplate \cite{jewett2013moltemplate}, a tool for constructing large regular molecules to be used in LAMMPS simulations. Our Moltemplate structure files were organized hierarchically, with: tubulin monomers arranged into $\alpha$-$\beta$ dimer pairs; which were then arranged into rings of thirteen dimers; which were then stacked to create a molecule 48 dimers long. Note that this organization has no effect on the final LAMMPS simulation: we report it here for reproducibility, as well as providing the template files in the supplementary material accompanying this paper. 

For this model, we define energetic interactions for angles and associations only. No steric or dihedral interactions were used: for dihedrals, this was because the lattice structure of the tube meant any set of four molecules contributed to multiple, contradictory dihedral interactions \footnote{Association and angle constraints were sufficient to replicate the bending resistance behavior of microtubules. We hope to run a similar experiment using higher-order particle interactions (which may be more biologically plausible), in future work.}. Interaction energy of an association $b$ was calculated using the ``harmonic'' bond style in LAMMPS, i.e. 
$
    E(b) = k{(\text{length}(b) - b_0)}^2,
$ where $b_0$ is the resting length and $k$ is the strength of that interaction. The energy of an angle $\phi$ was similarly calculated using the ``harmonic'' angle style, i.e.
$
E(\phi) = k{(\phi - \phi_0)}^2,
$ where $\phi_0$ is the resting angle and $k$ is again the interaction strength. The resting lengths and angles for all energetic interactions were calculated using the resting geometry of our microtubule graph $G_\text{mt}$: a LAMMPS script was used to print the value of every angle interaction in the model, and these were collected and grouped based on value (all $153^\circ$ angles, all $102\circ$ angles, etc). Each strength parameter was varied over the values in $\{.1,.3,.6, 1.0,1.3,1.6,1.9\}$, producing $7^5$ parameter combinations. Langevin dynamics were used, but with small temperature, to ensure stability and emphasize mechanical interactions.  See Table \ref{tab:ener_inter} and Figure \ref{fig:mt_labelled} for details on each strength parameter. See Figure \ref{fig:param_vary} for an illustration of varying resting positions and final energies as a result of varying these interaction parameters. 

\begin{figure}[h]
        \centering
        \includegraphics[width=\linewidth]{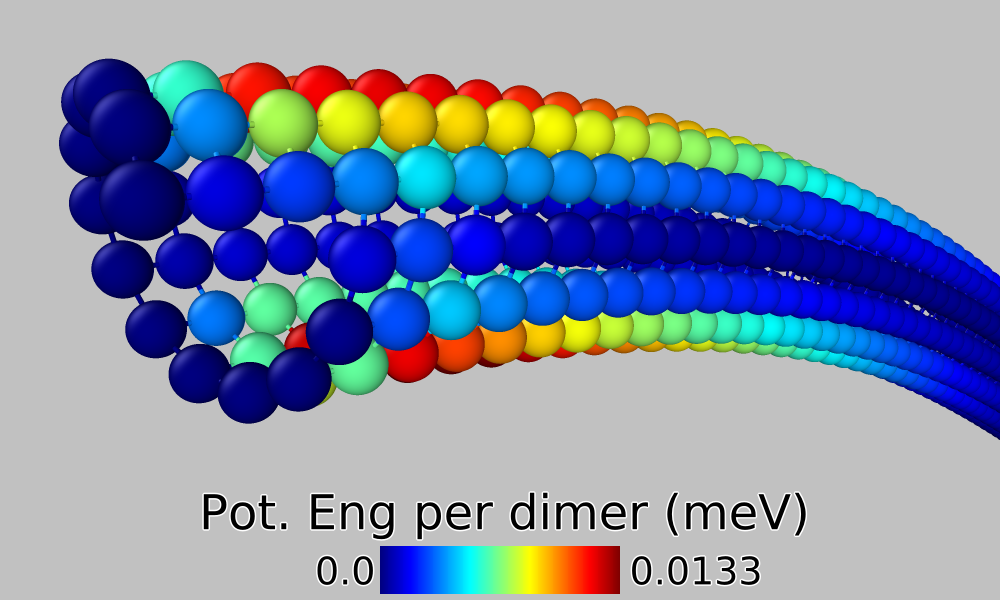}
        \caption{Microtubule model under bending load. Color of each particle indicates the sum of that particle's share of all of the energetic interactions in which it participates. This view is of the clamped end; the other end, out of view, has a constant force applied.}
        \label{fig:bend_mt}
    \end{figure}

\begin{figure}[h]
        \centering
        \includegraphics[width=\linewidth]{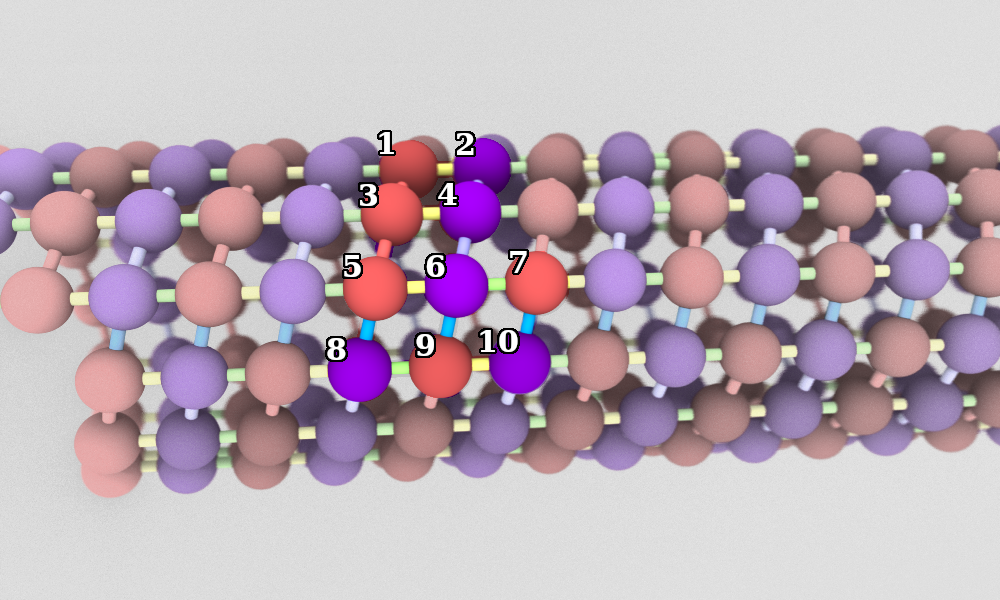}
        \caption{Microtubule model structure. Red spheres represent $\alpha$-tubulin; purple spheres represent $\beta$-tubulin. Highlighted atoms at center are labelled to show example energetic interactions: each type of interaction indicated in Table \ref{tab:ener_inter} (using the particle labels in this image) is applied everywhere in the model where that arrangement of particle and association types occurs in that position.}
        \label{fig:mt_labelled}
    \end{figure}

\begin{figure}
        \centering
        
        \includegraphics[width=.85\linewidth]{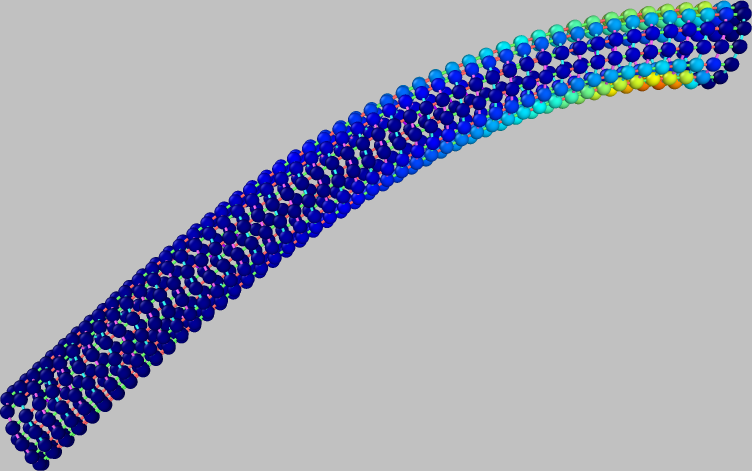} \\
        \vspace{5pt}
        \includegraphics[width=.85\linewidth]{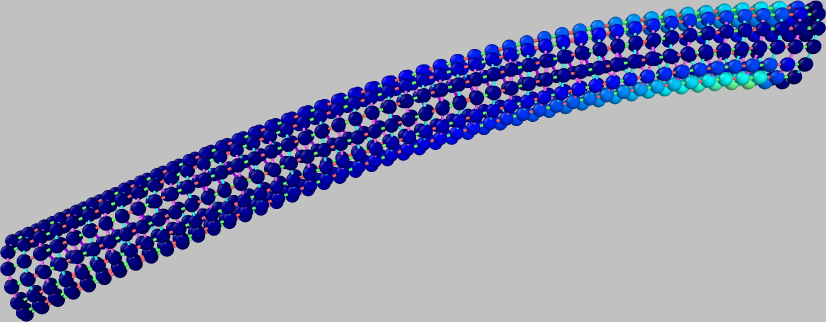} \\
        \vspace{5pt}
        
        \includegraphics[width=.85\linewidth]{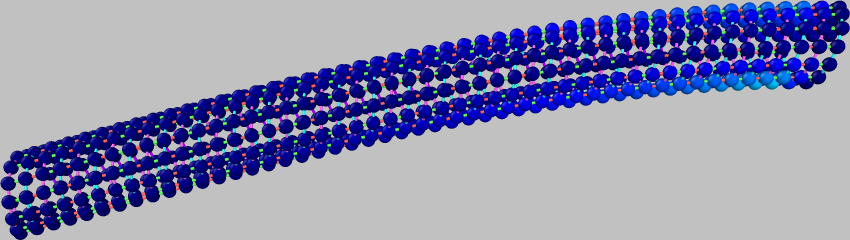} \\
        \vspace{5pt}
        
        \includegraphics[width=.85\linewidth]{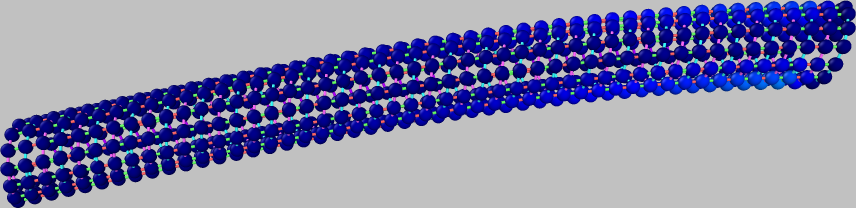} \\
        \vspace{5pt}
        
        \includegraphics[width=.85\linewidth]{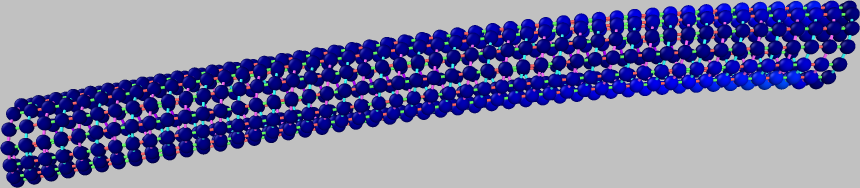} \\
        \vspace{5pt}
        
        \includegraphics[width=.85\linewidth]{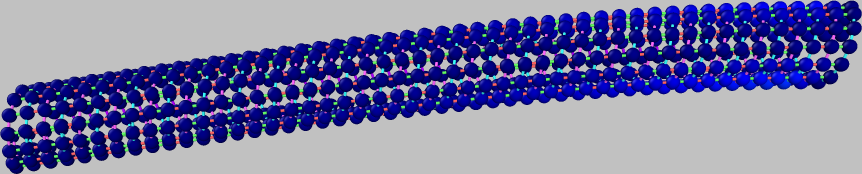} \\
        \vspace{5pt}
        
        \includegraphics[width=.85\linewidth]{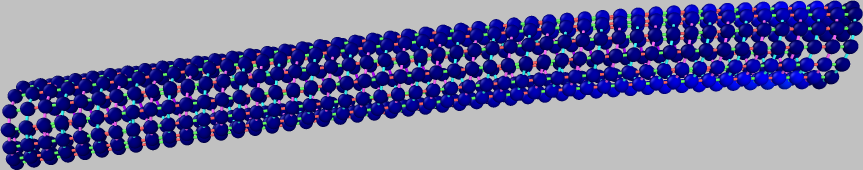} \\

        \caption{Changes in stiffness of microtubule model under constant load, as parameters controlling interaction strength are varied. From top to bottom, all parameters are set to the same values in $\{.1,.3,.6, 1.0,1.3,1.6,1.9\}$. Particles (tubulin monomers) are colored according to their contribution to total potential energy of the configuration, identically to Figure \ref{fig:bend_mt}. All pictures show the microtubule at rest e.g. at the end of the simulation run using that parameter set.}
        \label{fig:param_vary}
    \end{figure}

GNU Parallel \cite{Tange2011a} was used to run a simulation for each combination of interaction parameters, using the particle dynamics simulation engine LAMMPS. In each simulation, we clamp the first two rings of tubulin monomers (nodes 1-26) in place, and apply force (in the negative $y$ direction) to the final two rings of monomers (nodes 599-624). This force starts at 0 and ramps up during the first 128000 timesteps (one step $ = .5$ns) to its maximum value of $3 \times 10^{-15}$N. Once maximum force is reached, the simulation runs for 256000 additional timesteps, which in our experience was long enough for all particles to come to rest. See Figure \ref{fig:bend_mt} for an illustration (visualized with Ovito \cite{ovito}) of the potential energy per-particle at the final frame of a typical simulation run. Every $K=32000$ timesteps, we save the following for every particle: the position $x,y,z$; components of velocity $v_x, v_y, v_z$; components of force $F_x, F_y, F_z$; and the potential energy of the particle $E$. The dataset is then a concatenation of the 12 saved frames from every simulation run, comprising all combinations of input parameter values, where for each frame we have: \\
 $x_i$, the input graph signal, a $624 \times 10$ matrix holding the position and velocity of each particle, as well as values of the four interaction coefficients; and \\
$y_i$, the output graph signal, a $624 \times 1$ matrix holding the potential energy calculated for each particle.

    During training, after a training/validation split, we normalize the data by taking the mean and standard deviation of the $N_\text{train} \times 624 \times 10$ input and $N_\text{train} \times 624 \times 1$ output tensors along their first axis. Each data tensor is then reduced by the mean and divided by the standard deviation so that all  $624 \times 10$ inputs to the network have zero mean and unit standard deviation. We normalize using the training data only.

    \begin{table*}[t]
        \centering
        \caption{Description of energetic interactions in microtubule simulation, according to the labels in Figure \ref{fig:mt_labelled}.}
        \begin{small}
        \begin{center}
        \begin{tabular}{c}
        \toprule 
        \begin{sc}
        Association interactions
        \end{sc}  \\
        \midrule 
        \begin{tabular}{cccc}
        Description & Examples & Resting Length & Strength Param. \\
        Lateral association inside lattice & (1,3),(2,4)  & 5.15639nm & {\sc LatAssoc} \\
        Lateral association across seam & (5,8),(6,9)  & 5.15639nm & {\sc LatAssoc} \\
        Longitudinal association & (1,2),(3,4) & 5.0nm & {\sc LongAssoc} \\
        \end{tabular} \\
        \bottomrule
        \toprule 
        \begin{sc}
        angle interactions
        \end{sc} \\
        \midrule 
        \begin{tabular}{cccc}
        Description & Examples & Resting Angle & Strength Param. \\
        Pitch angle inside lattice & (1,3,5),(2,4,6)  & 153.023$^{\circ}$ & {\sc LatAngle} \\
        Longitudinal angle & (5,6,7),(8,9,10) & 180$^{\circ}$ & {\sc LongAngle} \\
        Lattice cell acute angle & (3,4,6),(3,5,6),(5,8,9),(6,9,10) & 77.0694$^{\circ}$ & {\sc QuadAngles} \\
        Lattice cell obtuse angle & (4,3,5),(4,6,5),(6,5,8),(6,9,8) & 102.931$^{\circ}$ & {\sc QuadAngles} 

        \end{tabular} \\
        \bottomrule
        \end{tabular}    
        \end{center}
        \end{small}
        \label{tab:ener_inter}
    \end{table*}

\subsection{Graph Coarsening}
\label{subsec:coarse}

\begin{figure}[h]
    \centering
    \includegraphics[width=\linewidth]{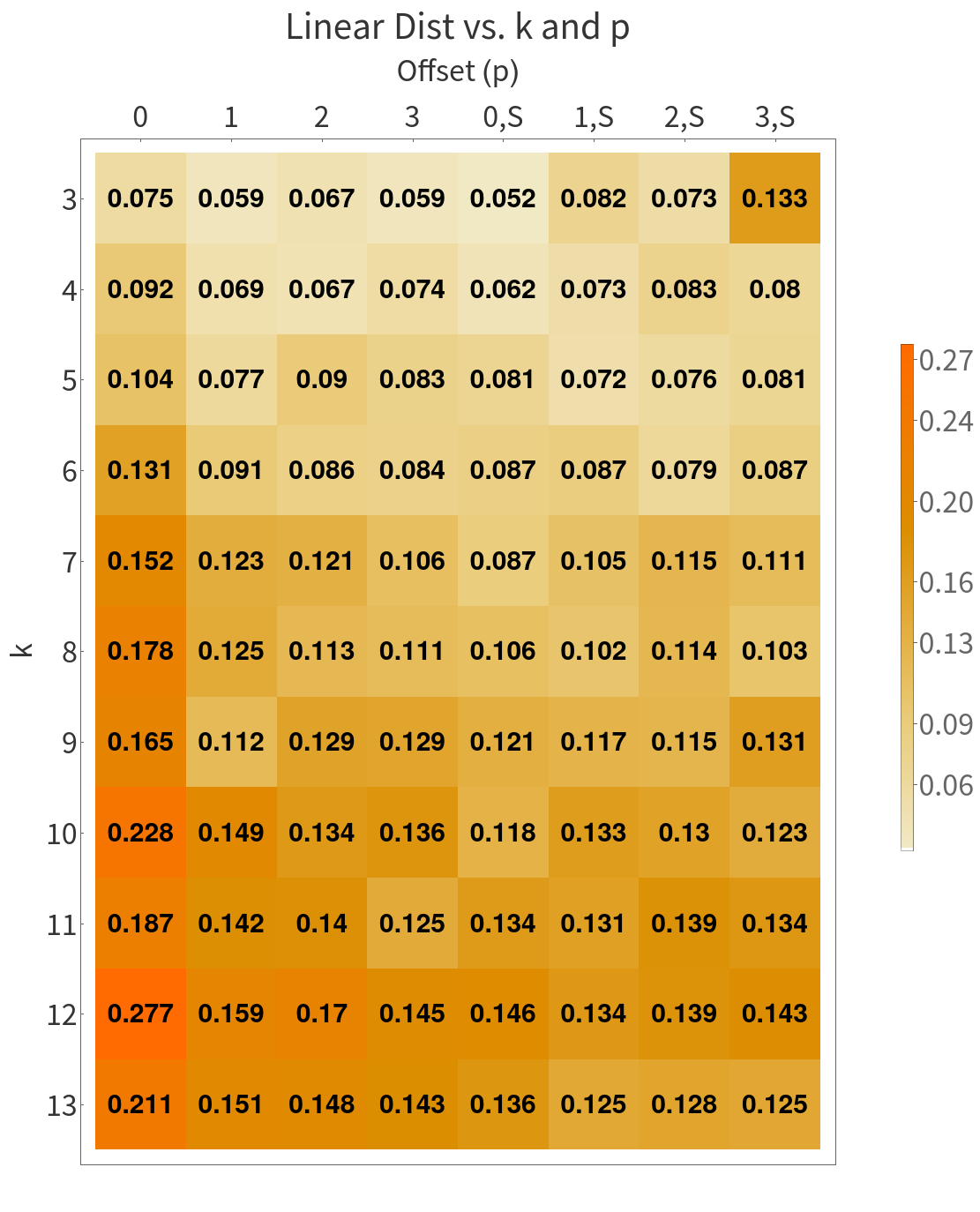}
    \caption{Directed Graph Diffusion Distance (GDD) between offset tube graphs and $G_\text{mt}$. Table cells colored by value. We see from this comparison that the two graphs which are closest to $G_\text{mt}$ are $G_{\text{Tube}(24,3,0)}$ and $G_{\text{Tube}(24,3,0)}$ with an edge weight of 2 for connections along the seam, motivating our choice of $G_{\text{Tube}(24,3,0)}$ (unweighted) as the coarsest graph in our hierarchy.}
    \label{fig:mt_dist_table}
\end{figure}

\begin{figure}[h]
    \centering
        \begin{tabular}{cc}
        $G_\text{mt}$ & \raisebox{-.5\height}{\includegraphics[width=.6\linewidth]{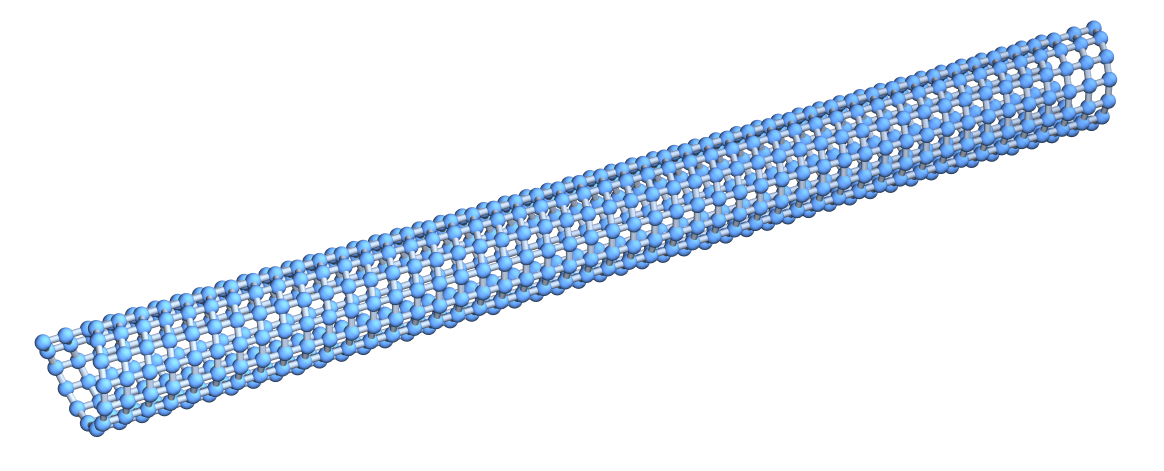}} 
        \end{tabular} \\
        \vspace{-.66cm}
        \begin{tabular}{cc}
        $G_\text{inter}$ & \raisebox{-.5\height}{\includegraphics[width=.6\linewidth]{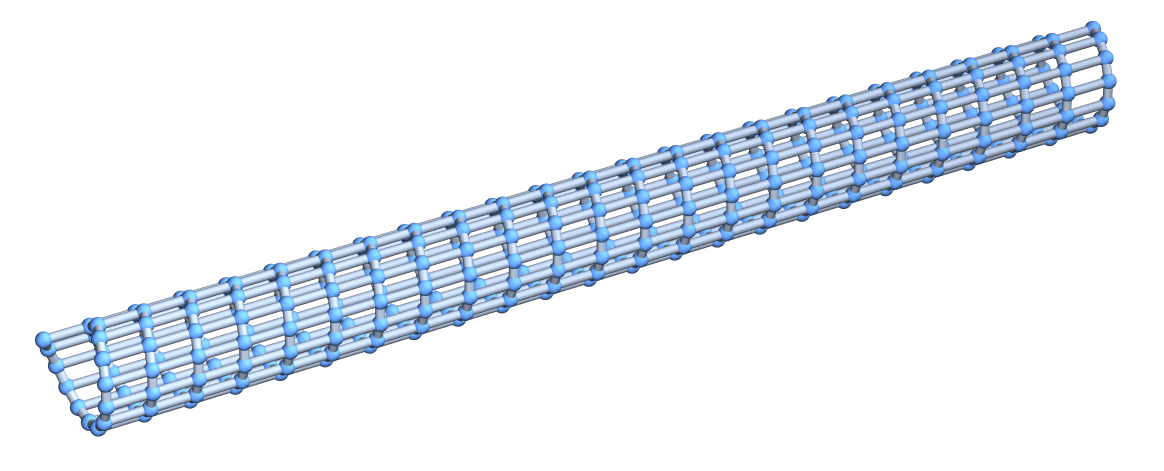}} 
        \end{tabular} \\
        \vspace{-.66cm}
        \begin{tabular}{cc}
        $G_\text{coarse}$ & \raisebox{-.5\height}{\includegraphics[width=.6\linewidth]{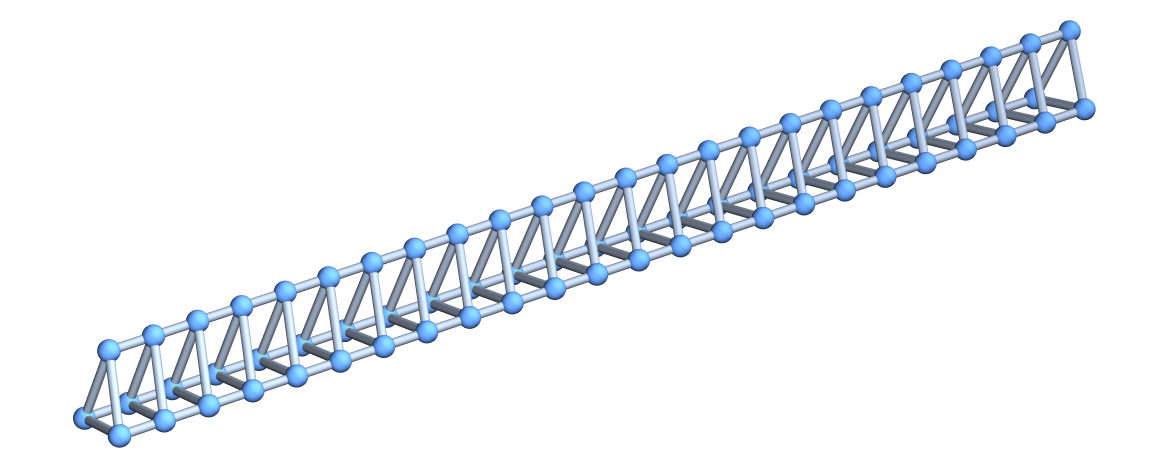}} \\
    \end{tabular}
    \caption{Three graphs used to create structure matrices for our GPCN model. Top: microtubule graph. Center: Offset tube with 13 subunits per turn, length 24, and offset 1. Bottom: Tube with 3 subunits per turn, no offset, and length 24.}
    \label{fig:my_label}
\end{figure}

In this Section we outline a procedure for determining the coarsened structure matrices to use in the hierarchy of GCN models comprising a GPCN. We use our microtubule graph as an example. In this case, we have two a-priori guidelines for producing the reduced-order graphs: 1) the reduced models should still be a tube and 2) it makes sense from a biological point of view to coarsen by combining the $\alpha$-$\beta$ pairs into single subunits. Given these restrictions, we can explore the space of coarsened graphs and find the coarse graph which is nearest to our original graph (under the GDD). 

Our microtubule model is a tube of length 48 units, 13 units per complete ``turn'', and with the seam offset by three units. We generalize this notion as follows: Let $p$ be the offset, and $k$ be the number of monomers in one turn of the tube, and $n$ the number of turns of a tube graph $G_{\text{Tube}(n,k,p)}$. The graph used in our simulation is thus $G_\text{mt} = G_{\text{Tube}(48,13,3)}$. We pick the medium scale model $G_\text{inter}$ to be $G_{\text{Tube}(24,13,1)}$, as this is the result of combining each $\alpha\text{-}\beta$ pair of tubulin monomer units in the fine scale, into one tubulin dimer unit in the medium scale. We pick the coarsest graph $G_\text{coarse}$ by searching over possible offset tube graphs. Namely, we vary $k \in \{3,4,\ldots 12\}$ and $p \in \{0,1,2,3\}$, and compute the optimal $P^*$ and its associated distance $D(G_\text{Tube(24,k,p)}, G_\text{mt} | P = P^*)$. Figure \ref{fig:mt_dist_table} shows the distance between $G_\text{mt}$ and various other tube graphs as parameters $p$ and $k$ are varied. The nearest $G_\text{Tube(24,k,p)}$ to $G_\text{mt}$ is that with $p=0$ and $k=3$. Note that Figure \ref{fig:mt_dist_table} has two columns for each value of $k$: these represent the coarse edges along the seam having weight (relative to the other edges) 1 (marked with an $S$) or having weight 2 (no $S$). This is motivated by the fact that our initial condensing of each dimer pair condensed pairs of seam edges into single edges.

\subsection{Comparison to Other GCN Ensemble Models}
\label{subsec:compare_gcn}

\begin{table}[h]
    \centering
    \begin{small}
    \caption{Filter specifications for ensemble models in comparison experiment.}
    \label{tab:ensemble_models}
    \begin{tabular}{c}
    \toprule
        \begin{tabular}{c|c|c}
             Structure Matrix & GCN Filters & Dense Filters\\
        \end{tabular} \\
    \toprule    
        Single GCN \\
        \midrule
        \begin{tabular}{c|c|c}
             $L_\text{mt}$ & 64,64,64 & 256, 32, 8, 1 \\
        \end{tabular} \\
    \toprule    
        2-GCN Ensemble \\
        \midrule
        \begin{tabular}{c|c|c}
             $L_\text{mt}$ & 64,64,64 & 256, 32, 8, 1 \\
             $L_\text{mt}$ & 32,32,32 & 256, 32, 8, 1 \\
        \end{tabular} \\
    \toprule   
        3-GCN Ensemble \\
        \midrule
        \begin{tabular}{c|c|c}
             $L_\text{mt}$ & 64,64,64 & 256, 32, 8, 1 \\
             $L_\text{mt}$ & 32,32,32 & 256, 32, 8, 1 \\
             $L_\text{mt}$ & 16,16,16 & 256, 32, 8, 1 \\
        \end{tabular} \\
    \toprule
        2-level GPCN \\
        \midrule
        \begin{tabular}{c|c|c}
             $L_\text{inter}$ & 64,64,64 & 256, 32, 8, 1 \\
             $L_\text{mt}$ & 32,32,32 & 256, 32, 8, 1 \\
        \end{tabular} \\
    \toprule
        3-level GPCN \\
        \midrule
        \begin{tabular}{c|c|c}
             $L_\text{coarse}$ & 64,64,64 & 256, 32, 8, 1 \\
             $L_\text{inter}$ & 32,32,32 & 256, 32, 8, 1 \\
             $L_\text{mt}$ & 16,16,16 & 256, 32, 8, 1 \\
        \end{tabular} \\
    \toprule    
        N-GCN (radii 1,2,4) \\
        \midrule
        \begin{tabular}{c|c|c}
             $L_\text{mt}^r$ & 64,64,64 & 256, 32, 8, 1 \\
        \end{tabular} \\
    \toprule        
        N-GCN (radii 1,2,4,8,16) \\
        \midrule
        \begin{tabular}{c|c|c}
             $L_\text{mt}^r$ & 64,64,64 & 256, 32, 8, 1 \\
        \end{tabular} \\
    \toprule    
    \end{tabular}
    \end{small}
\end{table}

\begin{figure}[h]
    \centering
    \includegraphics[width=.88\linewidth]{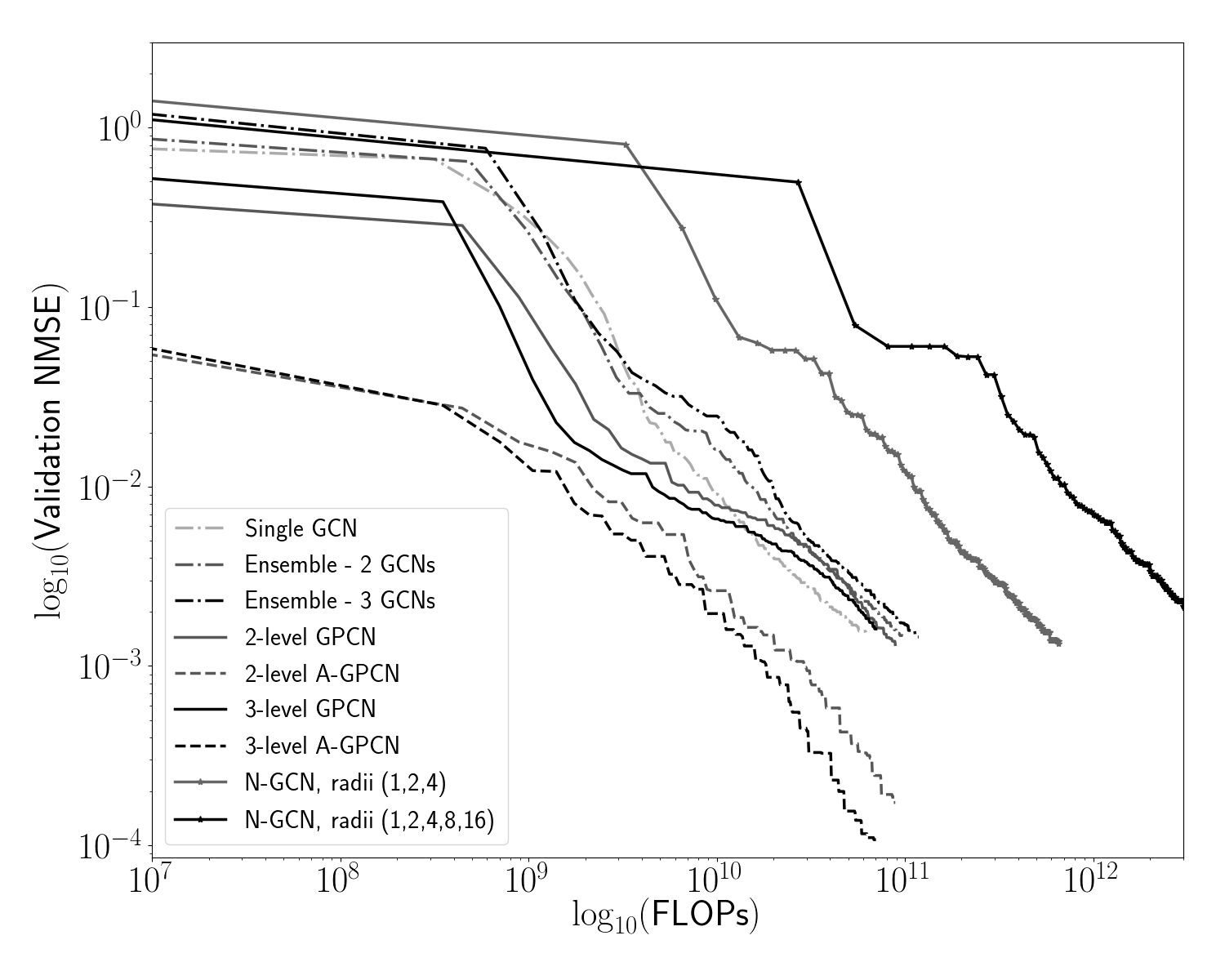}
    \caption{Comparison of Normalized MSE on held-out validation data as a function of FLOPs expended for a variety of ensemble Graph Convolutional Network Models. We see that especially in early stages of training, our model formulation learns faster than an ensemble of 2, 3 or 5 GCNs with the same number of filters. The error plotted is the model's minimum error thus far (on the validation data).}
    \label{fig:gcn_comp}
\end{figure}
\begin{table}[h]
    \caption{Mean error and uncertainty of several GCN ensemble models across ten random trials. For each trial, the random seed was set to the same value for each model. Reported values are the minimum error on the validation set during training (not the error at the final epoch). Normalized Mean Squared Error (NMSE) values are unitless. Only one trial was performed with the \textsc{DiffPool} model. }
    \vskip 0.15in
    \begin{center}
    \begin{small}
    \begin{tabular}{rcc}
        \toprule
        Model Name & 
        \begin{tabular}{c}
             Mean NMSE \\
             $\pm$ Std. Dev  \\
             ($\times 10^{-3}$)
        \end{tabular}
          &
        \begin{tabular}{c}
             Min NMSE \\
             ($\times 10^{-3}$)
        \end{tabular} \\
        \midrule
Single GCN & 1.55 $\pm$ 0.10 & 1.45914 \\
Ensemble - 2 GCNs & 1.44 $\pm$ 0.07 & 1.38313 \\
Ensemble - 3 GCNs & 1.71 $\pm$ 0.20 & 1.43059 \\
2-level GPCN & 1.43 $\pm$ 0.12 & 1.24838 \\
2-level A-GPCN & 0.17 $\pm$ 0.05 & 0.08963 \\
3-level GPCN & 2.09 $\pm$ 0.32 & 1.57199 \\
3-level A-GPCN & 0.131 $\pm$ 0.030 & 0.10148 \\
\begin{tabular}{r}
     N-GCN  \\
     radii (1,2,4)
\end{tabular} & 1.30 $\pm$ 0.05 & 1.23875 \\
\begin{tabular}{r}
     N-GCN  \\
     radii (1,2,4,8,16)
\end{tabular} & 1.30 $\pm$ 0.06 & 1.22023 \\
DiffPool & 2.041 $\pm$ n/a & 2.041 \\
        \bottomrule

    \end{tabular}
    \end{small}
    \end{center}
    \label{tab:err_comparison}
\end{table}

\begin{table}[h]
    \centering
    
    \caption{Mean wall-clock time to perform feed-forward and backpropagation for one batch of data, for various GCN ensemble models. Times were collected on a single Intel(R) Xeon(R) CPU core and an NVIDIA TITAN X GPU.}
    \vskip 0.15in
    \begin{small}
    \begin{tabular}{rl}
    \toprule
    Model Name & Mean time per batch (s)\\ 
    \midrule
        Single GCN & 0.042 \\
        Ensemble - 2 GCNs & 0.047 \\
        Ensemble - 3 GCNs & 0.056 \\
        2-level GPCN & 0.056 \\
        2-level A-GPCN & 0.056 \\
        3-level GPCN & 0.061 \\
        3-level A-GPCN & 0.059 \\
        N-GCN, radii (1,2,4) & 0.067 \\
        N-GCN, radii (1,2,4,8,16) & 0.086 \\
        DiffPool & 0.0934 \\
    \bottomrule
    \end{tabular}
    \end{small}
    \label{tab:wallclock}
\end{table}

In this experiment we demonstrate the efficiency advantages of our model by comparing our approach to other ensemble Graph Convolutional Networks. Within each ensemble, each GCN model consists of several graph convolution layers, followed by several dense layers which are applied to each node separately (node-wise dense layers can be alternatively understood as a GCN layer with $Z = I$, although we implement it differently for efficiency reasons). The input to the dense layers is the node-wise concatenation of the output of each GCN layer. Each ensemble is the sum output of several such GCNs. We compare our models to 1, 2, and 3- member GCN ensembles with the same number of filters (but all using the original fine-scale structure matrix). For GPCN models, $P$ matrices were calculated using Pymanopt \cite{townsend2016pymanopt} to optimize Equation \ref{eqn:GDD} subject to orthogonality constraints. The same $P$ were used to initialize the (variable) $P$ matrices of A-GPCN models.  

We also compare our model to the work of Abu-El-Haija et. al \yrcite{abuelhaija2018ngcn}, who introduce the N-GCN model: an ensemble GCN in which each ensemble member uses a different power $Z^r$ of the structure matrix (to aggregate information from neighborhoods of radius $r$). We include a N-GCN with radii (1,2,4) and a N-GCN with radii (1,2,4,8,16).

All models were trained with the same train/validation split, using ADAM with default hyperparameters, in TensorFlow \cite{abadi2016tensorflow}. Random seeds for Python, TensorFlow, Numpy, and Scipy were all initialized to the same value for each training run, to ensure that the train/validation split is the same across all experiments, and the batches of drawn data are the same. See supplementary material for version numbers of all software packages used. Training batch size was set to 8, all GCN layers have ReLU activation, and all dense layers have sigmoidal activation with the exception of the output layer of each network (which is linear). All modes were trained for 1000 epochs of 20 batches each. The time per batch of each model is listed in Table \ref{tab:wallclock}. 
Since hardware implementations may differ, we estimate the computational cost in FLOPs of each operation in our models. The cost of a graph convolutional layer with $n \times n$ structure matrix $Z$, $n \times F$ input data $X$, and $F \times C$ filter matrix $W$ is estimated as: $nF(|Z| + C)$, where $|Z|$ is the number of nonzero entries of $Z$. This is calculated as the sum of the costs of the two matrix multiplications $X \cdot W$ and $Z \cdot XW$, with the latter assumed to be implemented as sparse matrix multiplication and therefore requiring $O(|Z| n F)$ operations. For implementation reasons, our GCN layers (across all models) do not use sparse multiplication; if support for arbitrary-dimensional sparse tensor outer products is included in TensorFlow in the future, we would expect the wall-clock times in Table \ref{tab:wallclock} to decrease. The cost of a dense layer (with $n \times F$ input data $X$, and $F \times C$ filter matrix $W$) applied to every node separately is estimated as: $O(n F C)$. The cost of taking the dot product between a $n \times k$ matrix and a $k \times m$ matrix (for example, the restriction/prolongation by $P$) is estimated as $O(n m k)$. 

We summarize the structure of each of our models in Table \ref{tab:ensemble_models}.  In Figure \ref{fig:gcn_comp} we show a comparison between each of these models, for one particular random seed (42). Error on the validation set is tracked as a function of computational cost expended to train the model (under our cost assumption given above). We see that all four GPCN models outperform the other types of ensemble model during early training, in the sense that they reach lower levels of error for the same amount of computational work performed. Additionally, the adaptive GPCN models outperform all other models in terms of absolute error: after the same number of training epochs (using the same random seed) they reach an order of magnitude lower error. Table \ref{tab:err_comparison} shows summary statistics for several runs of this experiment with varying random seeds; we see that the A-GPCN models consistently outperform all other models considered. Note that Figures \ref{fig:gcn_comp},\ref{fig:diff_comp}, and \ref{fig:gpcn_schedule} plot the Normalize Mean Squared Error (NMSE). This unitless value compares the output signal to the target after both are normalized by the procedure described in section \ref{subsec:data_gen}.

\subsection{Comparison: All-at-Once or Coarse-to-Fine Training}
\label{subsec:compare_layerwise}
In this Section we compare the computational cost of training the entire GPCN at once, versus training the different `resolutions' (meaning the different GCNs in the hierarchy) of the network according to a more complicated training schedule. This approach is motivated by recent work in coarse-to-fine training of both flat and convolutional neural networks \cite{scott2019multilevel, zhao2019pgunet,haber2018learning,dou2015coarse,ke2017multigrid}, as well as the extensive literature on Algebraic MultiGrid (AMG) methods \cite{vanvek1996algebraic}.

AMG solvers for differential equations on a mesh (which arises as the discretization of some volume to be simulated) proceed by performing numerical ``smoothing steps'' at multiple resolutions of discretization. The intuition behind this approach is that modes of error should be smooth at a spatial scale which is equivalent to their wavelength, i.e. the solver shouldn't spend many cycles resolving long-wavelength errors at the finest scale, since they can be resolved more efficiently at the coarse scale. Given a solver and a hierarchy of discretizations, the AMG literature defines several types of training procedures or ``cycle'' types (F-cycle, V-cycle, W-cycle). These cycles can be understood as being specified by a recursion parameter $\gamma$, which controls how many times the smoothing or training algorithm visits all of the coarser levels of the hierarchy in between smoothing steps at a given scale. For example, when $\gamma = 1$ the algorithm proceeds from fine to coarse and back again, performing one smoothing step at each resolution - a `V' cycle. 

We investigate the efficiency of training 3-level GPCN and A-GPCN (as described in Section \ref{subsec:compare_gcn}), using multigrid-like training schedules with $\gamma \in \{ 0, 1, 2, 3\}$, as well as ``coarse-to-fine'' training: training the coarse model to convergence, then training the coarse and intermediate models together (until convergence), then finally training all three models at once. Error was calculated at the fine-scale. For coarse-to-fine training convergence was defined to have occurred once 10 epochs had passed without improvement of the validation error. 

\begin{figure}[h]
    \centering
    \includegraphics[width=.88\linewidth]{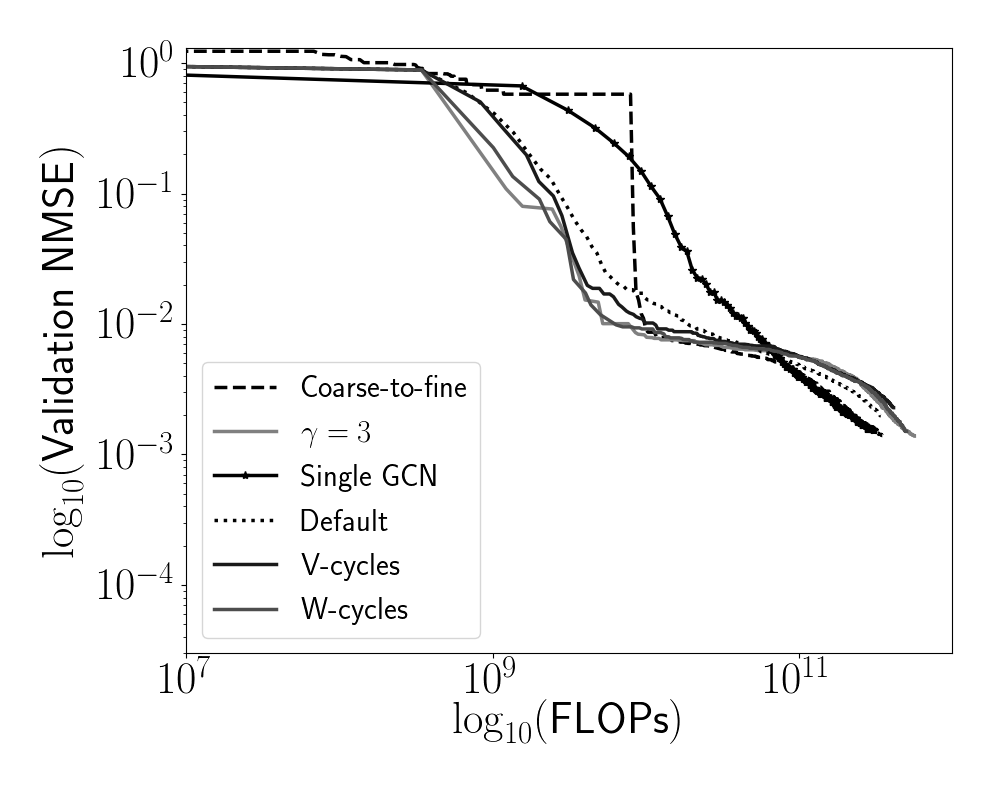} \\
    \includegraphics[width=.88\linewidth]{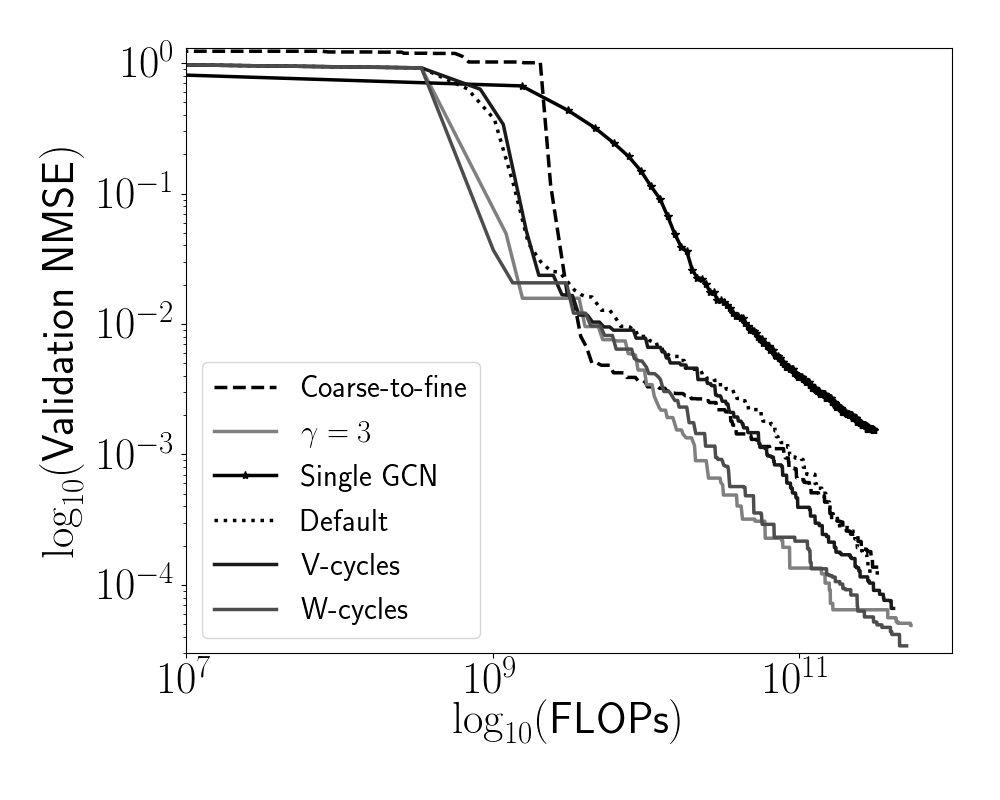}
    \caption{Effect of varying training schedule for training a GPCN model. Notably, The various multigrid training cycles result in models which are more accurate, and do so more efficiently. Top: FLOPs vs. NMSE for training GPCNs with multigrid training schedules. Bottom: same, but with A-GPCNs. }
    \label{fig:gpcn_schedule}
\end{figure}

\begin{figure}[h]
        \centering
        \includegraphics[width=.88\linewidth]{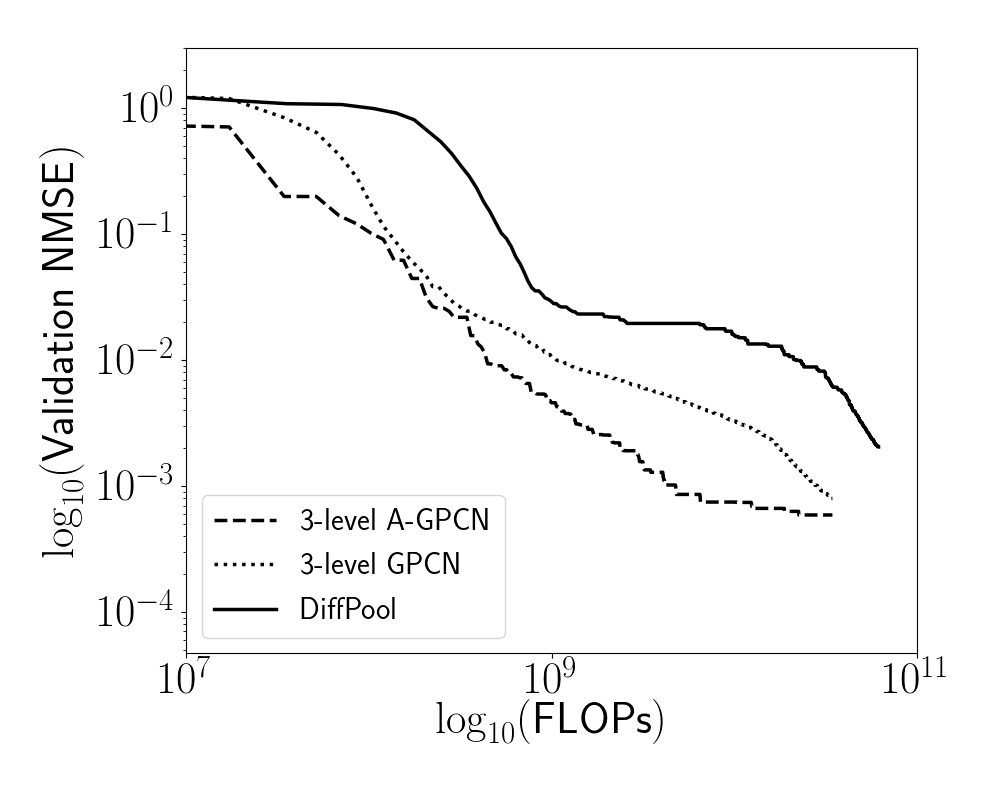}
        \caption{Comparison of 3-level GPCN and A-GPCN models to a 3-level GPCN which uses {\sc DiffPool} modules to coarsen the input graph and data. Our models improve over {\sc DiffPool} in terms of both efficiency and final error.}
        \label{fig:diff_comp}
    \end{figure}

Our experiments (see Figure \ref{fig:gpcn_schedule}) show that these training schedules do result in a slight increase in efficiency of the GPCN model, especially during the early phase of training. The increase is especially pronounced for the schedules with $\gamma=2$ and $\gamma=3$. Furthermore, these multigrid training schedules produce models which are more accurate than the GPCN and A-GPCN models trained in the default manner. 

\subsection{Comparison with DiffPool}

Graph coarsening procedures are in general not differentiable. DiffPool \cite{ying2018hierarchical} aims to address this by constructing an auxiliary GCN, whose output is a pooling matrix. Formally: 
 Suppose that at layer $l$ of a GCN we have a $n_l \times n_l$ structure matrix $Z^{(l)}$ and a $n \times F$ data matrix $X^{(l)}$. In addition to GCN layers as described in Section \ref{sec:model_info}, Ying et. al define a pooling operation at layer $l$ as:
        \begin{align*}
            S^{(l)} &= \sigma \left( \textsc{gcn}_\text{pool}\left(Z^{(l)}, X^{(l)}, {\left\{\theta^{(i)}_1 \right\}}_{l=1}^{m}\right) \right)
        \end{align*}
        where $\textsc{gcn}_\text{pool}$ is an auxillary GCN with its own set of parameters ${\left\{\theta^{(i)}_1 \right\}}_{l=1}^{m}$, and $\sigma$ is the softmax function. The output of $\textsc{gcn}_\text{pool}$ is a $n \times n_\text{coarse}$ matrix, each row of which is softmaxed to produce an affinity matrix $S$ whose rows each sum to 1, representing each fine-scale node being connected to one unit's worth of coarse-scale nodes. The coarsened structural and data matrices for the next layer are then calculated as:
        \begin{align}
            X^{(l+1)} &= {S^{(l)}}^T X^{(l)} \nonumber \\
            Z^{(l+1)} &= {S^{(l)}}^T Z^{(l)} {S^{(l)}} \label{eqn:diffpool_rule}
        \end{align}
        Clearly, the additional GCN layers required to produce ${S^{(l)}}$ incur additional computational cost. We compare our 3-level GPCN (adaptive and not) models from the experiment in Section \ref{subsec:compare_gcn} to a model which has the same structure, but in which each $P$ matrix is replaced by the appropriately-sized output of a {\sc DiffPool} module, and furthermore the coarsened structure matrices are produced as in Equation \ref{eqn:diffpool_rule}.  
        
        We see that our GPCN model achieves comparable validation loss with less computational work, and our A-GPCN model additionally achieves lower absolute validation loss.

\section{Future Work}
\label{sec:future}

\subsection{Differentiable Models of Molecular Dynamics}
This work demonstrates the use of feed-forward neural networks to approximate the energetic potentials of a mechanochemical model of an organic molecule. Per-timestep, GCN models may not be as fast as highly-parallelized, optimized MD codes. However, neural networks are highly flexible function approximators: the GCN training approach outlined in this paper could also be used to train a GCN which predicts the energy levels per particle at the end of a simulation (once equilibrium is reached), given the boundary conditions and initial conditions of each particle. In the case of our MT experiments, approximately $3 \times 10^5$ steps were required to reach equilibrium. The computational work to generate a suitably large and diverse training set would then be amortized by the GCN's ability to generalize to initial conditions, boundary conditions, and hyperparameters outside of this
data set.
Furthermore, this GCN reduced model would be fully differentiable, making it possible to perform gradient descent with respect to any of these inputs. 
In particular, we derive here the gradient of the input to a GCN model with respect to its inputs. 
\subsubsection{Derivation of Energy Gradient {w.r.t} Position}
As described above, the output of our GCN (or GPCN) model is a $n \times 1$ matrix (or vector) $Y$. The total energy of the molecule at position $X$ is given by $E = \sum_{i=1}^n \left[Y\right]_i$. Note that any GCN's initial layer update is given by the update rule:
     \begin{align*}
        X^` &= g_1\left(Z X W_1 +b_1 \right).
     \end{align*}
     During backpropagation, as an intermediate step of computing the partial derivatives of energy with respect to $W_1$ and $b_1$, we must compute the partial $\frac{\partial E}{\partial A_1}$ of energy with respect to the input to the activation function $g_1$:
     \begin{align*}
         A_1 &= Z X W_1 +b_1 \\
         X^` &= g_1(A_1). \\
     \end{align*}
     We therefore assume we have this derivative. By the Chain Rule for matrix derivatives:
     \begin{align}
         {\left[\frac{\partial E}{\partial X} \right]}_{ij} = \frac{\partial E}{\partial {\left[ X \right]}_{ij} } &= \sum_{k,p} \frac{\partial E}{\partial {\left[ A_1 \right]}_{kp}} \frac{\partial {\left[ A_1 \right]}_{kp}}{\partial [x_{ij}]}. \nonumber
         \intertext{Since}
         {\left[ A_1 \right]}_{kp} &= \left( \sum_{c,d} {\left[ Z \right]}_{kc}{\left[ X \right]}_{cd}{\left[ W_1 \right]}_{dp} \right) + {\left[ b_1 \right]}_{kp} \nonumber
         \intertext{ and therefore }
         \frac{\partial {\left[ A_1 \right]}_{kp}}{\partial {\left[ X \right]}_{ij}} &= {\left[ Z \right]}_{ki}{\left[ W_1 \right]}_{jp}, \nonumber \\
         \frac{\partial E}{\partial {\left[ X \right]}_{ij}} &= \sum_{k,p} \frac{\partial E}{\partial {\left[ A_1 \right]}_{kp}} {\left[ Z \right]}_{ki} {\left[ W_1 \right]}_{jp} \nonumber \\
         \frac{\partial E}{\partial X} &= Z^T \frac{\partial E}{\partial A_1} W_1^T. \label{eqn:gnn_backprop_rule}
     \end{align}

Furthermore, since our GPCN model is a sum of the output of several GCNs, we can also derive a backpropagation equation for the gradient of the fine-scale input, $\mathbf{X}$, with respect to the energy prediction of the entire ensemble. Let $E^{(i)}$ represent the fine-scale energy prediction of the $i$th member of the ensemble, so that $E = \sum_{i=1}^k E^{(i)}$. Then, let  
\begin{align}
    \frac{\partial E^{(i)}}{\partial X^{(i)}} &= {Z^{(i)}}^T \frac{\partial E^{(i)}}{\partial A^{(i)}_1} {W^{(i)}_1}^T 
\end{align}
be the application of Equation \ref{eqn:gnn_backprop_rule} to each GCN in the ensemble.
Since the input to the $i$th member of the ensemble is given by $X^{(i)} = P_{1,i}^T \mathbf{X}$, we can calculate the gradient of $E^{(i)}$ with respect to $\mathbf{X}$, again using the Chain Rule:

\begin{align}
    \frac{\partial E^{(i)}}{\partial {\left[ \mathbf{X} \right]}_{mn} } &= \sum_{s=1}^{N_s} \sum_{t=1}^{N_t} \frac{\partial E^{(i)}}{\partial {\left[ X^{(i)} \right]}_{st} } \frac{\partial {\left[ X^{(i)} \right]}_{st} }{\partial {\left[ \mathbf{X} \right]}_{mn}} \nonumber \\
     &= \sum_{s=1}^{N_s} \sum_{t=1}^{N_t} \frac{\partial E^{(i)}}{\partial {\left[ X^{(i)} \right]}_{st}} \frac{\partial \left[ P_{1,i}^T \mathbf{X} \right]_{st}}{\partial {\left[ \mathbf{X} \right]}_{mn}} \nonumber \\ 
     &= \sum_{s=1}^{N_s} \sum_{t=1}^{N_t} \frac{\partial E^{(i)}}{\partial {\left[ X^{(i)} \right]}_{st}} \delta_{tm} {\left[ P_{1,i} \right]}_{ns} \nonumber \\ 
     &= \sum_{s=1}^{N_s} \frac{\partial E^{(i)}}{\partial {\left[ X^{(i)} \right]}_{sm}} {\left[ P_{1,i} \right]}_{ns} \nonumber \\ 
    \intertext{Therefore,}
    \frac{\partial E^{(i)}}{\partial {\left[ \mathbf{X} \right]}_{mn}} &= P_{1,i} \frac{\partial E^{(i)}}{\partial X^{(i)}} \nonumber
    \intertext{and so}
    \frac{\partial E}{\partial \mathbf{X}} &=  \sum_{i=1}^k \frac{\partial E^{(i)}}{\partial \mathbf{X}} = \sum_{i=1}^k P_{1,i} \frac{\partial E^{(i)}}{\partial X^{(i)}} \nonumber
\end{align}

This backpropagation rule may then be used to adjust $\mathbf{X}$, and thereby find low-energy configurations of the molecular graph. Additionally, analogous to the GCN training procedure outlined in Section \ref{subsec:compare_layerwise}, this optimization over molecule positions could start at the coarse scale and be gradually refined. 

\subsection{Tensor Factorization}
Recent work has re-examined GCNs in the context of the extensive literature on tensor decompositions. LanczosNet \cite{liao2019lanczosnet}, uses QR decomposition of the structure matrix to aggregate information from large neighborhoods of the graph. The ``Tensor Graph Convolutional Network" of Zhang et. al \yrcite{zhang2018tensor}, is a different decomposition method, based on graph factorization; a product of GCNs operating on each factor graph can be as accurate as a single GCN acting on the product graph. Since recent work \cite{scott2019multilevel} has shown that the GDD of a graph product is bounded by the distances between the factor graphs, it seems reasonable to combine both ideas into a model which uses a separate GPCN for each factor. One major benefit of this approach would be that a transfer-learning style approach can be used. For example, we could train a product of two GCN models on a short section of microtubule; and then re-use the weights in a model that predicts energetic potentials for a longer microtubule. This would allow us to extend our approach to MT models whose lengths are biologically relevant, e.g. $10^3$ tubulin monomers.

\subsection{Graph Limits}
\label{subsec:graph_limits}
Given that \emph{in vivo} microtubules are longer than the one simulated in this paper by a factor of as much as 200x, future work will focus on scaling these methods to the limit of very large graphs. In particular, this means repeating the experiments of Sections \ref{sec:exp}, but with longer tube graphs. We hypothesise that tube graphs which are closer to the microtubule graph (under the LGDD) as their length $n \rightarrow \infty$ will be more efficient reduced-order models for a GPCN hierarchy. This idea is similar to the ``graphons'' (which are the limits of sequences of graphs which are Cauchy under the Cut-Distance of graphs) introduced by  Lov{\'a}sz \cite{lovasz2012large}. To show that it is reasonable to define a ``graph limit'' of microtubule graphs in this way, we plot the distance between successively longer microtubule graphs. Using the same notation as in Section \ref{subsec:coarse}, we define three families of graphs:
\begin{itemize}
    \item $G_{\text{Grid}}(n,13)$: Grids of dimensions $n \times 13$, and;
    \item $G_{\text{Tube}(n,13,1)}$: Microtubule graphs with $13$ protofilaments, of length $n$, with offset 1, and;
    \item $G_{\text{Tube}(2n,13,3)}$: Microtubule graphs with $13$ protofilaments, of length $2n$, with offset 3.
\end{itemize}
In this preliminary example, as $n$ is increased, we see a clear distinction in the distances $D(G_{\text{Tube}(n,13,1)},G_{\text{Tube}(2n,13,3)})$ and $D(G_{\text{Grid}(n,13)},G_{\text{Tube}(2n,13,3)})$, with the former clearly limiting to a larger value as $n \rightarrow \infty$.
\begin{figure}
    \centering
    \includegraphics[width=\linewidth]{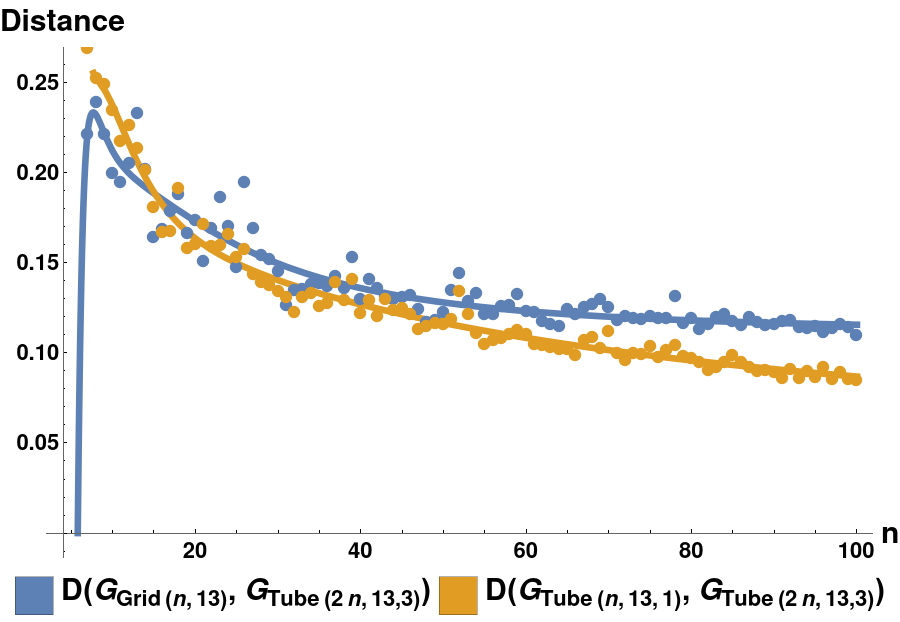}
    \caption{Limiting behavior of two classes of distances between graphs, as a function of graph size. We plot $D(G_{\text{Tube}(n,13,1)},G_{\text{Tube}(2n,13,3)})$ and
     $D(G_{\text{Grid}(n,13)},G_{\text{Tube}(2n,13,3)})$ as a function of $n$, along with seventh-degree polynomial fit curves of each. The smaller tube graphs are closer than the grid graphs to the larger tube, even in the large-graph limit. 
    }
    \label{fig:limit_test}
\end{figure}
\section{Conclusion}
\label{sec:conclude}
We introduce a new type of graph ensemble model which explicitly learns to approximate behavior at multiple levels of coarsening. Our model outperforms several other types of GCN, including both other ensemble models and a model which coarsens the original graph using DiffPool. We also explore the effect of various training schedules, discovering that A-GPCNs can be effectively trained using a coarse-to-fine training schedule. We present the first use of GCNs to approximate energetic potentials in a model of a microtubule.

\section*{Acknowledgements}

\ifanon
Funding\footnote{Grant numbers redacted for anonymity.} provided by from the U.S. National Institute of Aging grant \censor{AG059602},
Human Frontiers Science Program grant HFSP - \censor{RGP0023/2018}, U.S. National Science Foundation NRT Award number \censor{1633631}, and the Leverhulme Trust.
\else
Funding provided by from the U.S. National Institute of Aging grant AG059602,
Human Frontiers Science Program grant HFSP - RGP0023/2018, U.S. National Science Foundation NRT Award number 1633631, and the Leverhulme Trust.
\fi

\bibliography{refs}

\begin{thebibliography}{31}
\providecommand{\natexlab}[1]{#1}
\providecommand{\url}[1]{\texttt{#1}}
\expandafter\ifx\csname urlstyle\endcsname\relax
  \providecommand{\doi}[1]{doi: #1}\else
  \providecommand{\doi}{doi: \begingroup \urlstyle{rm}\Url}\fi

\bibitem[Abadi et~al.(2016)]{abadi2016tensorflow}
Abadi, M. et~al.
\newblock {T}ensorflow: A {S}ystem for {L}arge-{S}cale {M}achine {L}earning.
\newblock In \emph{12th $\{$USENIX$\}$ Symposium on Operating Systems Design
  and Implementation ($\{$OSDI$\}$ 16)}, pp.\  265--283, 2016.

\bibitem[Abu-El-Haija et~al.(2018)Abu-El-Haija, Kapoor, Perozzi, and
  Lee]{abuelhaija2018ngcn}
Abu-El-Haija, S., Kapoor, A., Perozzi, B., and Lee, J.
\newblock {N-GCN}: {M}ulti-{S}cale {G}raph {C}onvolution for
  {S}emi-{s}upervised {N}ode {C}lassification, 2018.

\bibitem[{B}acciu et~al.(2019){B}acciu, {E}rrica, {M}icheli, and
  {P}odda]{bacciu2019gentle}
{B}acciu, D., {E}rrica, F., {M}icheli, A., and {P}odda, M.
\newblock {A} {G}entle {I}ntroduction to {D}eep {L}earning for {G}raphs.
\newblock \emph{arXiv preprint arXiv:1912.12693}, 2019.

\bibitem[Bijsterbosch \& Volgenant(2010)Bijsterbosch and
  Volgenant]{bijsterbosch2010solving}
Bijsterbosch, J. and Volgenant, A.
\newblock Solving the rectangular assignment problem and applications.
\newblock \emph{Annals of Operations Research}, 181\penalty0 (1):\penalty0
  443--462, 2010.

\bibitem[Chakrabortty et~al.(2018)Chakrabortty, Blilou, Scheres, and
  Mulder]{chakrabortty2018computational}
Chakrabortty, B., Blilou, I., Scheres, B., and Mulder, B.~M.
\newblock {A} {C}omputational {F}ramework for {C}ortical {M}icrotubule
  {D}ynamics in {R}ealistically {S}haped {P}lant {C}ells.
\newblock \emph{PLoS Computational Biology}, 14\penalty0 (2):\penalty0
  e1005959, 2018.

\bibitem[Dou \& Wu(2015)Dou and Wu]{dou2015coarse}
Dou, H. and Wu, X.
\newblock {C}oarse-to-{F}ine {T}rained {M}ulti-{S}cale {C}onvolutional {N}eural
  {N}etworks for {I}mage {C}lassification.
\newblock In \emph{2015 International Joint Conference on Neural Networks
  (IJCNN)}, pp.\  1--7. IEEE, 2015.

\bibitem[Gardner et~al.(2013)Gardner, Zanic, and
  Howard]{gardner2013microtubule}
Gardner, M.~K., Zanic, M., and Howard, J.
\newblock {M}icrotubule {C}atastrophe and {R}escue.
\newblock \emph{Current Opinion in Cell Biology}, 25\penalty0 (1):\penalty0
  14--22, 2013.

\bibitem[Haber et~al.(2018)Haber, Ruthotto, Holtham, and
  Jun]{haber2018learning}
Haber, E., Ruthotto, L., Holtham, E., and Jun, S.-H.
\newblock {L}earning {A}cross {S}cales - {M}ultiscale {M}ethods for
  {C}onvolution {N}eural {N}etworks.
\newblock In \emph{Thirty-Second AAAI Conference on Artificial Intelligence},
  2018.

\bibitem[Heindl(2018)]{heindl2018lapsolver}
Heindl, C.
\newblock {lapsolver}: {F}ast {L}inear {A}ssignment {P}roblem {(LAP)} {S}olvers
  for {P}ython {B}ased on {c}-{e}xtensions.
\newblock \url{https://github.com/cheind/py-lapsolver}, 2018.

\bibitem[Jewett et~al.(2013)Jewett, Zhuang, and Shea]{jewett2013moltemplate}
Jewett, A.~I., Zhuang, Z., and Shea, J.-E.
\newblock {M}oltemplate: a {C}oarse-{G}rained {M}odel {A}ssembly {T}ool.
\newblock \emph{Biophysical Journal}, 104\penalty0 (2):\penalty0 169a, 2013.

\bibitem[Ke et~al.(2017)Ke, Maire, and Yu]{ke2017multigrid}
Ke, T.-W., Maire, M., and Yu, S.~X.
\newblock {M}ultigrid {N}eural {A}rchitectures.
\newblock In \emph{Proceedings of the IEEE Conference on Computer Vision and
  Pattern Recognition}, pp.\  6665--6673, 2017.

\bibitem[{K}ipf \& {W}elling(2016){K}ipf and {W}elling]{kipf2016semi}
{K}ipf, T.~N. and {W}elling, M.
\newblock {S}emi-{S}upervised {C}lassification with {G}raph {C}onvolutional
  {N}etworks.
\newblock \emph{arXiv preprint arXiv:1609.02907}, 2016.

\bibitem[Kis et~al.(2002)Kis, Kasas, Babi{\'c}, Kulik, Benoit, Briggs,
  Sch{\"o}nenberger, Catsicas, and Forro]{kis2002nanomechanics}
Kis, A., Kasas, S., Babi{\'c}, B., Kulik, A., Benoit, W., Briggs, G.,
  Sch{\"o}nenberger, C., Catsicas, S., and Forro, L.
\newblock {N}anomechanics of {M}icrotubules.
\newblock \emph{Physical Review Letters}, 89\penalty0 (24):\penalty0 248101,
  2002.

\bibitem[Liao et~al.(2019)Liao, Zhao, Urtasun, and Zemel]{liao2019lanczosnet}
Liao, R., Zhao, Z., Urtasun, R., and Zemel, R.~S.
\newblock {L}anczos{N}et: {M}ulti-{S}cale {D}eep {G}raph {C}onvolutional
  {N}etworks.
\newblock \emph{arXiv preprint arXiv:1901.01484}, 2019.

\bibitem[Lov{\'a}sz(2012)]{lovasz2012large}
Lov{\'a}sz, L.
\newblock \emph{Large networks and graph limits}, volume~60.
\newblock American Mathematical Soc., 2012.

\bibitem[Pampaloni \& Florin(2008)Pampaloni and
  Florin]{pampaloni2008microtubule}
Pampaloni, F. and Florin, E.-L.
\newblock {M}icrotubule {A}rchitecture: {I}nspiration for {N}ovel {C}arbon
  {N}anotube-based {B}iomimetic {M}aterials.
\newblock \emph{Trends in Biotechnology}, 26\penalty0 (6):\penalty0 302--310,
  2008.

\bibitem[{P}limpton(1993)]{plimpton1993fast}
{P}limpton, S.
\newblock {F}ast {P}arallel {A}lgorithms {F}or {S}hort-{R}ange {M}olecular
  {D}ynamics.
\newblock Technical report, {S}andia {N}ational {L}abs., {A}lbuquerque, {NM}
  ({U}nited {S}tates), 1993.

\bibitem[Schneider \& Stoll(1978)Schneider and Stoll]{schneider1978molecular}
Schneider, T. and Stoll, E.
\newblock {M}olecular-{D}ynamics {S}tudy of a {T}hree-{D}imensional
  {O}ne-{C}omponent {M}odel for {D}istortive {P}hase {T}ransitions.
\newblock \emph{Physical Review B}, 17\penalty0 (3):\penalty0 1302, 1978.

\bibitem[Scott \& Mjolsness(2019{\natexlab{a}})Scott and
  Mjolsness]{scott2019multilevel}
Scott, C. and Mjolsness, E.
\newblock {M}ultilevel {A}rtificial {N}eural {N}etwork {T}raining for
  {S}patially {C}orrelated {L}earning.
\newblock \emph{{SIAM} {J}ournal on {S}cientific {C}omputing}, 41\penalty0
  (5):\penalty0 S297--S320, 2019{\natexlab{a}}.

\bibitem[Scott \& Mjolsness(2019{\natexlab{b}})Scott and
  Mjolsness]{scott2019novel}
Scott, C.~B. and Mjolsness, E.
\newblock Novel diffusion-derived distance measures for graphs,
  2019{\natexlab{b}}.

\bibitem[Shaw et~al.(2003)Shaw, Kamyar, and Ehrhardt]{shaw2003sustained}
Shaw, S.~L., Kamyar, R., and Ehrhardt, D.~W.
\newblock {S}ustained {M}icrotubule {T}readmilling in {A}rabidopsis {C}ortical
  {A}rrays.
\newblock \emph{Science}, 300\penalty0 (5626):\penalty0 1715--1718, 2003.

\bibitem[{S}tukowski({2010})]{ovito}
{S}tukowski, A.
\newblock {V}isualization and {A}nalysis of {A}tomistic {S}imulation {D}ata
  with {OVITO} - the {O}pen {V}isualization {T}ool.
\newblock \emph{{M}odelling {S}imulation in {M}aterials {S}cience and
  {E}ngineering}, {18}\penalty0 ({1}), {JAN} {2010}.
\newblock \doi{{10.1088/0965-0393/18/1/015012}}.

\bibitem[Tange(2011)]{Tange2011a}
Tange, O.
\newblock {GNU} {P}arallel - {T}he {C}ommand-{L}ine {P}ower {T}ool.
\newblock \emph{;login: The USENIX Magazine}, 36\penalty0 (1):\penalty0 42--47,
  Feb 2011.
\newblock \doi{http://dx.doi.org/10.5281/zenodo.16303}.
\newblock URL \url{http://www.gnu.org/s/parallel}.

\bibitem[Tindemans et~al.(2014)Tindemans, Deinum, Lindeboom, and
  Mulder]{tindemans2014efficient}
Tindemans, S.~H., Deinum, E.~E., Lindeboom, J.~J., and Mulder, B.
\newblock {E}fficient {E}vent-{D}riven {S}imulations {S}hed {N}ew {L}ight on
  {M}icrotubule {O}rganization in the {P}lant {C}ortical {A}rray.
\newblock \emph{Frontiers in Physics}, 2:\penalty0 19, 2014.

\bibitem[Townsend et~al.(2016)Townsend, Koep, and
  Weichwald]{townsend2016pymanopt}
Townsend, J., Koep, N., and Weichwald, S.
\newblock {P}ymanopt: A {p}ython {T}oolbox for {O}ptimization on {M}anifolds
  using {A}utomatic {D}ifferentiation.
\newblock \emph{The {J}ournal of {M}achine {L}earning {R}esearch}, 17\penalty0
  (1):\penalty0 4755--4759, 2016.

\bibitem[VanBuren et~al.(2005)VanBuren, Cassimeris, and
  Odde]{vanburen2005mechanochemical}
VanBuren, V., Cassimeris, L., and Odde, D.~J.
\newblock {M}echanochemical {M}odel of {M}icrotubule {S}tructure and
  {S}elf-{A}ssembly {K}inetics.
\newblock \emph{Biophysical Journal}, 89\penalty0 (5):\penalty0 2911--2926,
  2005.

\bibitem[Van{\v{e}}k et~al.(1996)Van{\v{e}}k, Mandel, and
  Brezina]{vanvek1996algebraic}
Van{\v{e}}k, P., Mandel, J., and Brezina, M.
\newblock {A}lgebraic {M}ultigrid by {S}moothed {A}ggregation for {S}econd and
  {F}ourth {O}rder {E}lliptic {P}roblems.
\newblock \emph{{C}omputing}, 56\penalty0 (3):\penalty0 179--196, 1996.

\bibitem[Verlet(1967)]{verlet1967computer}
Verlet, L.
\newblock {C}omputer ``{E}xperiments" on {C}lassical {F}luids. {I}.
  {T}hermodynamical {P}roperties of {L}ennard-{J}ones {M}olecules.
\newblock \emph{{P}hysical {R}eview}, 159\penalty0 (1):\penalty0 98, 1967.

\bibitem[Ying et~al.(2018)Ying, You, Morris, Ren, Hamilton, and
  Leskovec]{ying2018hierarchical}
Ying, Z., You, J., Morris, C., Ren, X., Hamilton, W., and Leskovec, J.
\newblock {H}ierarchical {G}raph {R}epresentation {L}earning with
  {D}ifferentiable {P}ooling.
\newblock In \emph{Advances in Neural Information Processing Systems}, pp.\
  4800--4810, 2018.

\bibitem[Zhang et~al.(2018)Zhang, Zheng, Cui, and Li]{zhang2018tensor}
Zhang, T., Zheng, W., Cui, Z., and Li, Y.
\newblock {T}ensor {G}raph {C}onvolutional {N}eural {N}etwork.
\newblock \emph{arXiv preprint arXiv:1803.10071}, 2018.

\bibitem[Zhao et~al.(2019)Zhao, Dai, Zhang, Yu, Li, Li, Wang, and
  Zhang]{zhao2019pgunet}
Zhao, J., Dai, L., Zhang, M., Yu, F., Li, M., Li, H., Wang, W., and Zhang, L.
\newblock {PGU-net+}: {P}rogressive {G}rowing of {U-net+} for {A}utomated
  {C}ervical {N}uclei {S}egmentation.
\newblock \emph{Lecture Notes in Computer Science}, pp.\  51–58, Dec 2019.

\end{thebibliography}
\bibliographystyle{icml2020}

\end{document}


\twocolumn[
\icmltitle{SUPPLEMENTARY MATERIAL for:\\ ``Graph Prolongation Convolutional Networks'' 
           }

\icmlsetsymbol{equal}{*}

\begin{icmlauthorlist}
\icmlauthor{Cory B. Scott}{uci}
\icmlauthor{Eric Mjolsness}{uci}
\end{icmlauthorlist}

\icmlaffiliation{uci}{Department of Computer Science, University of California Irvine, Irvine, California, USA}

\icmlcorrespondingauthor{Cory B. Scott}{scottcb@uci.edu}

\icmlkeywords{Machine Learning, ICML}

\vskip 0.3in
]

\printAffiliationsAndNotice{} 

\section{Code}
\section{Figures}